\documentclass[conference]{IEEEtran}
\usepackage{cite}
\ifCLASSINFOpdf
  \usepackage[pdftex]{graphicx}
\else
  \usepackage[dvips]{graphicx}
\fi
\usepackage{stfloats}
\usepackage{url}

\usepackage[usenames,dvipsnames]{color}
\usepackage{setspace}
\usepackage{hyperref}
\usepackage{caption}
\usepackage{subcaption}
\usepackage{verbatim}
\usepackage{longtable}
\usepackage{multirow}
\usepackage{tabularx}

\newcommand{\todo}[1]{\textcolor{red}{TODO: #1}}

\newcommand{\sig}{\textsuperscript{\ddag}}
\newcommand{\nearsig}{\textsuperscript{\dag}}
\newcommand{\ssig}{\textsuperscript{++}}
\newcommand{\nnearsig}{\textsuperscript{+}}

\renewcommand{\baselinestretch}{0.936}

\begin{document}
\title{Analyzing the Language of Food on Social Media}

\author{
\IEEEauthorblockN{Daniel Fried, Mihai Surdeanu, Stephen Kobourov, Melanie Hingle, and Dane Bell}
\IEEEauthorblockA{University of Arizona, Tucson, AZ, USA\\
Email: \{dfried, msurdeanu, kobourov, hinglem, dane\}@email.arizona.edu
}
}

\maketitle

\begin{abstract}

We investigate the predictive power behind the language of food on
social media. We collect a corpus of over three million food-related
posts from Twitter and demonstrate that many latent population
characteristics can be directly predicted from this data: overweight
rate, diabetes rate, political leaning, and home geographical location of authors.
For all tasks, our language-based models significantly outperform the majority-class baselines. 
Performance is further improved with more complex natural language processing, such as topic modeling. 
We analyze which textual features have most predictive power for these
datasets, providing insight into the connections between the language
of food, geographic locale, and community characteristics. Lastly, we
design and implement an online system for real-time query and
visualization of the dataset. Visualization tools, such as 
geo-referenced heatmaps, semantics-preserving wordclouds and
temporal histograms, allow us to discover more complex, global patterns mirrored in
the language of food.
\end{abstract}
\IEEEpeerreviewmaketitle

\section{Introduction}
Our diets reflect our identities.
The food we eat is influenced by our lifestyles, habits, upbringing, cultural and family heritage.
In addition to reflecting our current selves, our diets also shape who we will be, by impacting our health and well-being.
The purpose of this work is to understand if information about individuals' diets, reflected in the language they use to describe their food, can convey latent information about a community, such as its location, likelihood of diabetes, and even political preferences. 
This information can be used for a variety of purposes, ranging from improving public health to better targeted marketing. 

In this work we use Twitter as a source of language about food.
The informal, colloquial nature of Twitter posts, as well as the ease of data access, 
make it possible to assemble a large corpus describing the type of food consumed and the context of the discussion.
\fnbelowfloat
\begin{figure}[b!]
    \centering
    \includegraphics[width=0.8\linewidth]{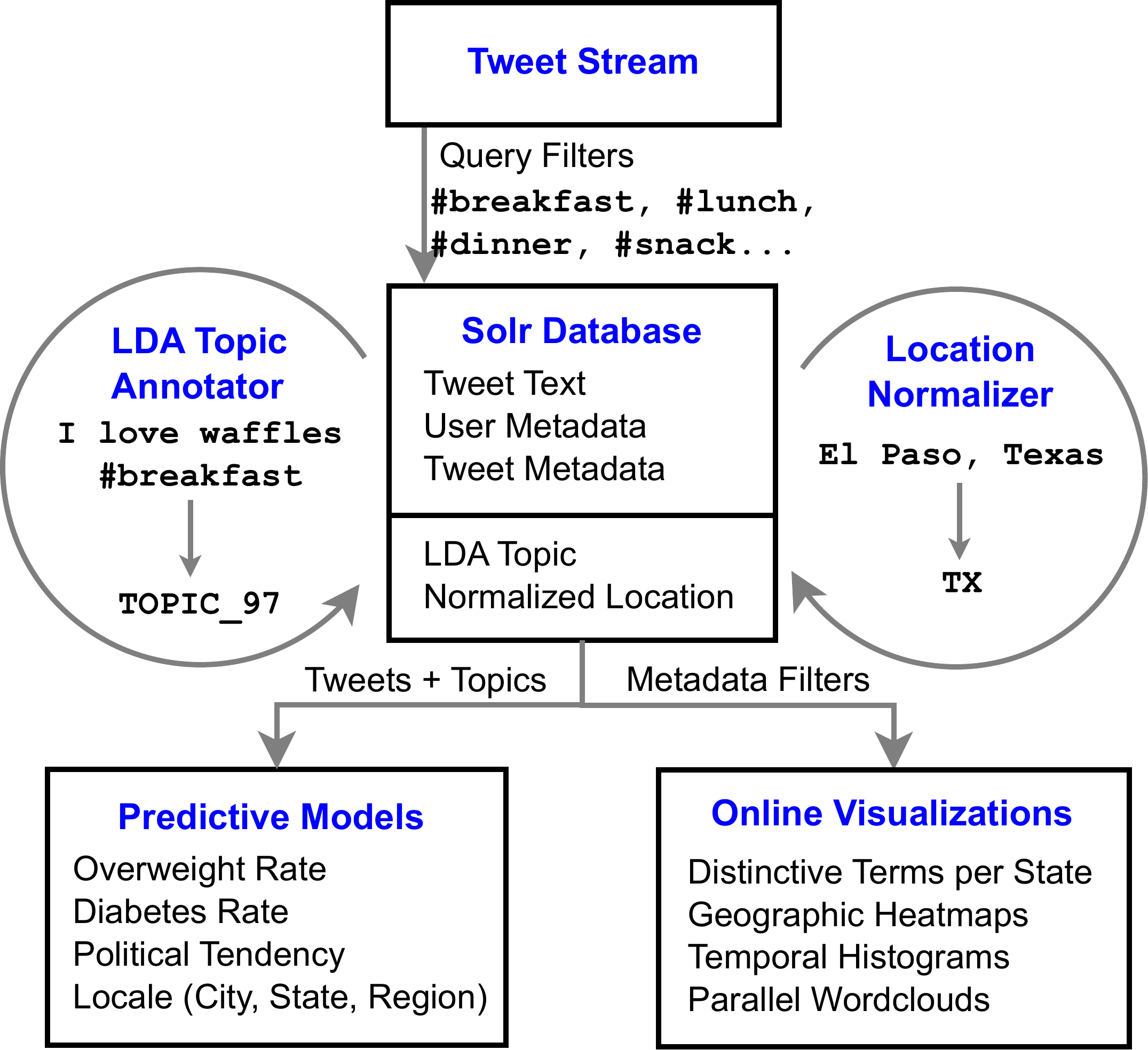}
\caption{\small \label{fig:system_diagram}The main steps of the system are: collecting tweets from Twitter using a set of meal-related filters, loading the tweets and their meta data into a Lucene-backed Solr instance, annotating the tweets with topic model labels (Section~\ref{sec:topical_features}) and normalizing locations (Section~\ref{data}), and then querying the tweets for use in the predictive models (Section~\ref{sec:prediction_tasks}) or visualization systems (Section~\ref{viz}).}
\end{figure}
Over eight months, we collected such a corpus of meal-related tweets together with relevant meta data, such as geographic locations and time of posting. We construct a system for aggregating, annotating, and querying these tweets as a source for predictive models and interactive visualizations (Fig.~\ref{fig:system_diagram}). Building on this dataset and system, the contributions of this work are fourfold:

\noindent \textbf{1.} We analyze the predictive power of the language of food by predicting several latent population characteristics from the tweets alone (after filtering out location-related words to avoid learning trivial correlations). We demonstrate that this data can be used to predict multiple characteristics, which are conceivably connected with food: a state's percentage of overweight population, the rate of diagnosed diabetes, and even political voting history.
Our results indicate that the language-based model yields statistically-significant improvements over the majority-class baseline in all configurations, and that more complex natural language processing (NLP), such as topic modeling, further improves results.

\noindent \textbf{2.} We demonstrate that the same data accurately predicts geographic home locale of the authors (from city-level, through state-level, to region-level), with our model significantly outperforming the random baseline (e.g., 30 times better on the state-level).

\noindent \textbf{3.} In addition to examining the effectiveness of our models on these predictive tasks, we analyze which textual features have most predictive power for these datasets, providing insight into the connections between the language of food, geographic locale, and community characteristics.

\noindent \textbf{4.} Lastly, we show that visualizations of the language of food over geographical or temporal dimensions can be used to infer additional information such as the importance of various daily meals in different regions, the distribution of different foods and drinks over the course of days, weeks and seasons, as well as some migration patterns in the United States and worldwide.

\section{Data}
\label{data}

Twitter provides an accessible source of data with broad demographic penetration across ethnicities, genders, and income levels\footnote{\url{http://www.pewinternet.org/2014/01/08/social-media-update-2013/twitter-users/}}, making it well-suited for examining the dietary habits of individuals on a large scale.
To identify and collect tweets about food, we queried Twitter's public streaming API\footnote{\url{https://dev.twitter.com/docs/api/streaming}. Note: Twitter caps the number of possible tweets returned by the streaming API to a fraction of the total number of tweets available at a given moment.} for posts containing hashtags related to meals (Table~\ref{tbl:hashtags}).
We collected approximately 3.5 million tweets containing at least one of these hashtags from the period between October 2, 2013 and May 29, 2014. 

Tweets are very short texts, limited to 140 characters. In our collection, the average length of a tweet is 8.7 words, after filtering out usernames, non-alphanumeric characters (hashtags excepted), and punctuation. The tweet collection contains a total of about 30 million words. Of these, there are around 1.5 million unique words.

\begin{table}[t]
    \small
    \centering
    \begin{tabular}{l|r|r}
        \hline
        \multirow{2}{*}{Term} & \multirow{2}{*}{\# of Tweets}& \# with normalized \\
             & & US Location \\
        \hline
        \texttt{\#dinner} & 1,156,630 & 173,634 \\
        \texttt{\#breakfast} & 979,031 & 161,214 \\
        \texttt{\#lunch} & 931,633 & 129,853 \\
        \texttt{\#brunch} & 287,305 & 86,239 \\
        \texttt{\#snack} & 139,136 & 21,539 \\
        \texttt{\#meal} & 94,266 & 12,149 \\
        \texttt{\#supper} & 32,235 & 2,971 \\
        \hline
        Total & 3,498,749 & 562,547\\
        \hline
    \end{tabular}
    \caption{\small \label{tbl:hashtags}Hashtags used to collect tweets, and number of tweets containing each hashtag. ``Normalized US location'' indicates that we could extract at least the user's state from the meta data. Since some tweets contain multiple meal hashtags, the total number of tweets (bottom row) is less than the column sum.}
\end{table}

Fig.~\ref{fig:system_diagram} describes the system used to collect, annotate, and process the tweets for prediction and visualization. Along with the text of each tweet, we store the user's self-reported location, time zone, and geotagging information, whenever these fields are available. 
This meta data is used to group tweets by the home location of the author, e.g., specified as city and/or state for those users located within the United States (US). For most experiments in this paper, geolocation normalization is performed using regular expressions, matching state names or postal abbreviations of one of the 50 US states or Washington, D.C. (e.g., {\em Texas} or {\em TX}), followed by matching city names or known abbreviations (e.g., {\em New York City} or {\em NYC}) within the author's location field. In case of ambiguities (e.g., {\em LA} stands for both Los Angeles and Louisiana) we used the user's time zone to disambiguate. About 16\% (562,547) of the collected tweets could be located within a state using this method (Table~\ref{tbl:hashtags}). 
We chose to use the self-reported user location instead of the geotagging information because: (a) it is more common, (b) it tends to have a standard, easily parseable form for US addresses, and (c) to avoid potential biases introduced by travel.
However, in Section~\ref{viz}, we extend our analysis to discover world-wide food-related patterns. In this context, because world addresses are considerably harder to parse than US addresses, we revert to geotagging information to identify the location of tweet authors.

\begin{figure}[t]
\begin{center}
    \includegraphics[width=\linewidth]{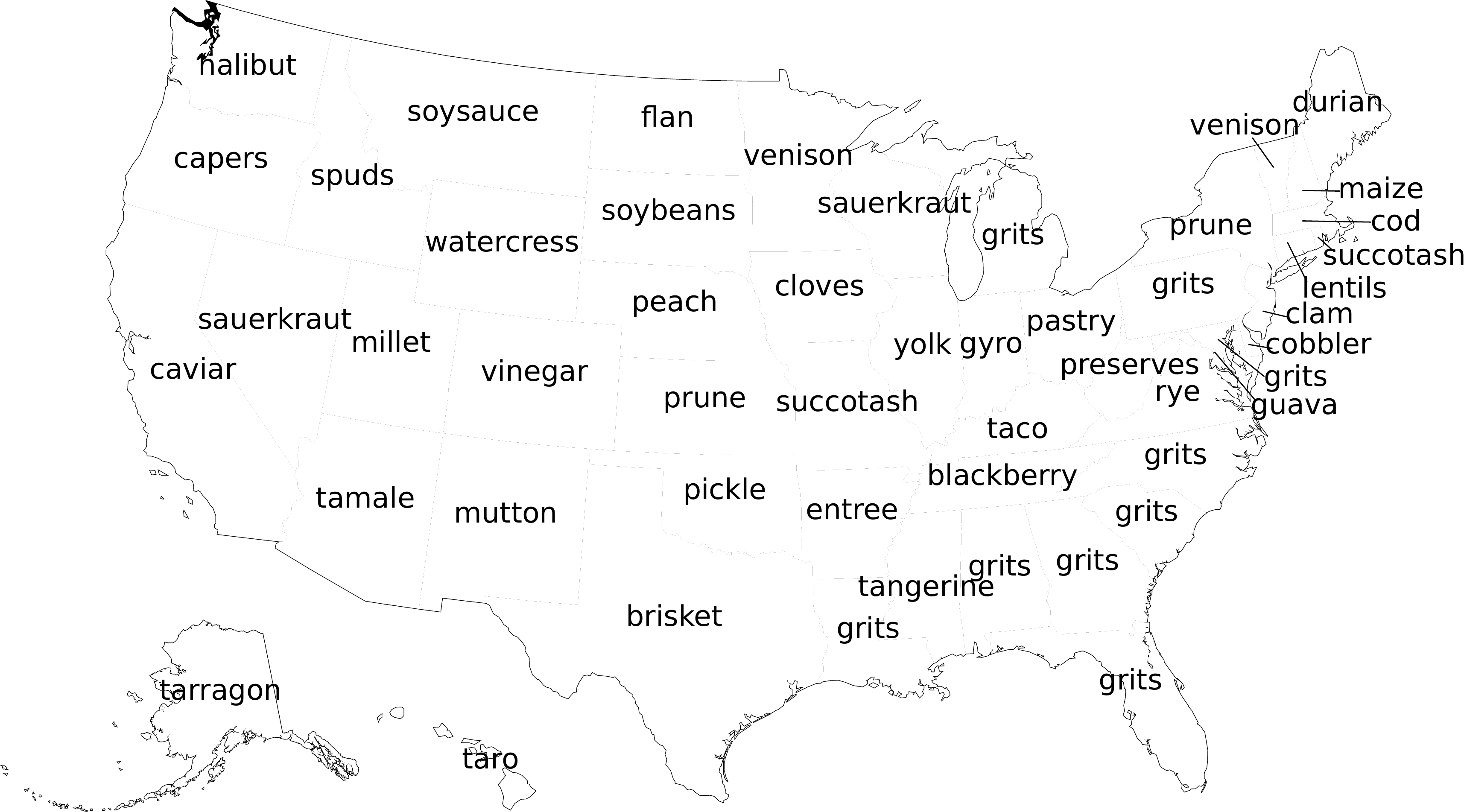}
\end{center}
\caption{\small \label{fig:terms_per_state}Most distinctive food word per state from the corpus of food-related tweets. Terms are filtered by a list containing about 800 food-related terms (Section~\ref{sec:lexical_features}) and ranked using {\em tf-idf}. Note that ``Prune'' is the name of a popular restaurant.}
\end{figure}

Using this dataset, we can immediately see food-driven patterns. For example, Fig.~\ref{fig:terms_per_state} shows prominent food-related words that appeared in the tweets normalized to each state. Tweet text is filtered using a list of approximately 800 food-related words (see Sec.~\ref{sec:lexical_features}). Terms are ranked using term frequency--inverse document frequency ({\em tf-idf})~\cite{manning2008iir} to discount words that occur frequently across all states, and give priority to those words that are highly representative of a state. Each state's food word with the highest {\em tf-idf} ranking is displayed in the map. Regional trends can be seen, for example {\em grits}, a breakfast food made from ground corn, is a common dish in the southern states, and various types of seafood ({\em halibut}, {\em caviar}, {\em cod}, {\em clam}) are popular in the eastern and western coastal states.

\section{Tasks}
\label{sec:prediction_tasks}

To understand the predictive power of the language of food, we implement several prediction tasks that use the tweets in the above dataset as their only input. We group these tasks into two categories: state-level characteristic prediction and locale prediction.

\subsection{Predicting State-Level Characteristics}
\label{sec:state_level}

Here we predict three aggregate characteristics for US states, using features extracted from the tweets produced by individuals in each state:

\subsubsection{Diabetes Rate} 
This is the percentage of adults in each state who have been told by a doctor that they have diabetes.
 Data in this set is taken from the Kaiser Commission on Medicaid and the Uninsured (KCMU)'s analysis of the Center for Disease Control's Behavioral Risk Factor Surveillance System (BRFSS) 2012 survey.\footnote{The BRFSS is a random-digit-dialed telephone survey of adults age 18 and over. For more details see: \url{http://kff.org/other/state-indicator/adults-with-diabetes/}} 
We convert this data into a binary dependent variable by considering whether a state's rate of diabetes is above or below the national median.
The median diabetes rate is 9.7\%, and the range is 6.0\% (7.0\% in Alaska to 13.0\% in West Virginia).
For example, Alabama has a diabetes rate of 12.3\%, which is above the national median of 9.7\%, so it is labelled as high-diabetes, while Alaska, with a rate of 7.0\%, is labelled as low-diabetes.

\subsubsection{Overweight Rate} 
This is the percentage of adults within each state who reported having a Body Mass Index (BMI) of at least 25.0 kilograms per meter squared, placing them within the ``overweight'' or ``obese'' categories defined by the National Institutes of Health.\footnote{\url{http://www.nhlbi.nih.gov/guidelines/obesity/BMI/bmi-m.htm}} 
As with the diabetes rate dataset, data is taken from KCMU's analysis of the BRFSS 2012 survey results\footnote{\url{http://kff.org/other/state-indicator/adult-overweightobesity-rate/}}.
Similarly, the corresponding binary dependent variable indicates if a state's overweight rate is above/below the national median.
The median overweight rate is 64.2\%, and the range is 17.7\% (51.9\% in Washington, D.C. to 69.6\% in Louisiana).

\subsubsection{Political Tendency} 
This dataset measures historical voting history  over a 5-year period: whether a state is more Democratic or Republican relative to the median US state, as measured by proportion of Democratic/Republican votes in general presidential, gubernatorial, and senatorial elections, in the interval from 2008 to 2013.\footnote{\url{http://uselectionatlas.org/}}. 
For example, Alaska cast 554,565 total votes for Democratic candidates and 748,488 for Republican candidates in these three types of elections during the six-year period, for a fraction of 42.6\% Democratic votes. This is below the median fraction of 51.6\%, so Alaska is labelled as \emph{Republican}.
Votes are compared relative to the median because of a slight bias toward Democratic votes during this time period.
The median fraction of Democratic votes is 51.6\%, and the range is 65.4\% (27.0\% in Wyoming to 92.4\% in Washington D.C.).

Because the above dependent variables are at state level, each state is treated as a single instance for these three tasks: all of the tweets produced within the state are aggregated into a single pool for feature extraction (detailed in the next section). We used Support Vector Machines (SVM) with a linear kernel~\cite{vapnik1998} for classification.

Although such a prediction task has many features (from all tweets in a given state), it has a small number of data points (51, one for each state plus Washington, D.C.). For this reason, we use leave-one-out cross-validation to evaluate the accuracy of the model. For each of the three data sets (overweight, diabetes, and political), we use the following process: Each state is held out in turn. The SVM is trained on features of tweets taken from the remaining 50 states, using the labels of the current data set. The SVM is then used to predict the current dataset's label of the held-out state. 
The accuracy of the model on the label set is calculated as the number of correct predictions out of the total number of states. To avoid overfitting, we do not tune  the classifier's hyper-parameters. 

\subsection{Predicting Locales}

To examine the connection between the language of food and geographic location, we seek to predict the locale of a group of tweets, using only the text of the tweets as input. 
We predict locales at different levels: city, state, and region.
It is important to note that, to focus our analysis on the predictive power of the language of food, we remove as many state and city names as possible from the tweets to avoid learning trivial correlations (see Sec.~\ref{sec:lexical_features}).

\subsubsection{City}
The locales in the city prediction task are the 15 most populous cities in the US.\footnote{\url{http://en.wikipedia.org/wiki/List_of_United_States_cities_by_population}} 

\subsubsection{State}
Locales in the state prediction task are the 50 US states, plus Washington, D.C. 
As discussed in the previous section, both city and state labels are assigned to tweets by parsing the self-reported author home location in the meta data.

\subsubsection{Region}
\label{sec:task_locale_region}
The final variant of the locale prediction task is to predict the geographic region of the US containing the user's state. We use four geographic regions, taken from the US Census Bureau: Midwest, West, Northeast, and South.\footnote{\url{http://www.census.gov/geo/maps-data/maps/pdfs/reference/us_regdiv.pdf}} 

Similar to the previous tasks, here we also aggregate tweets with the same locale for feature extraction, and use a linear kernel SVM for classification. However, the overall setup is different. 
Because the goal here is to predict locale itself, we divide the tweets from each locale into training and testing sets consisting of 80\% and 20\% of the tweets available for that locale, respectively. Tweets are sorted chronologically so that all training tweets for a given locale were posted before all the corresponding testing tweets. 
The overall training/testing corpora include training/testing tweets from all locales.
 Using this dataset, we train a one-against-many multi-class SVM classifier for the locale as the dependent variable. For example, for the city detection task, the SVM classifies a group of tweets as belonging to one of the 15 known cities. 

The accuracy on each locale prediction task is the number of locales correctly identified, divided by the total number of locales. Thus, a baseline classifier that randomly predicts a locale would achieve an accuracy of $1/51 = 1.96\%$ on the state prediction task, $1/15 = 7\%$ on the city task, and $1/4 = 25\%$ for region prediction.

\section{Feature Descriptions}

We use two sets of features: {\em lexical} (from tweet words) and {\em topical} (sets of words appearing in similar contexts). %

\subsection{Lexical Features}
\label{sec:lexical_features}

We take the simple approach of representing each locale as a bag of words assembled from all the tweets in that group. Each word becomes a feature with value equal to the number of times it occurrs across all tweets for that locale. We tokenize the tweets using the Stanford CoreNLP software.\footnote{\url{http://nlp.stanford.edu/software/corenlp.shtml}} An additional pre-processing step removes the following tokens: (a) tokens that do not contain alpha-numeric characters or punctuation (to reduce noise); (b) stopwords and words that occur a single time (to reduce data size); and, most importantly,  (c) URLs, usernames (preceded by an \texttt{@} symbol), and words and hashtags naming state and city locations\footnote{In addition of known state names and abbreviations we used a list of the 250 most populous cities in the US from \url{http://en.wikipedia.org/wiki/List_of_United_States_cities_by_population}, together with common nicknames, such as ``\#sanfran'' for San Francisco.} (to avoid learning trivial correlations, such as {\em \#TX} indicating a tweet from Texas).

We also experiment with open versus closed vocabularies. For open vocabularies, we use two configurations: all words produced by the above pre-processing step, or only hashtags. For a closed vocabulary experiment, we use a set of 809 words related to food, meals, and eating, obtained from the English portion of a Spanish-English food glossary\footnote{\url{http://www.lingolex.com/spanishfood/a-b.htm}} and an online food vocabulary list\footnote{\url{http://www.enchantedlearning.com/wordlist/food.shtml}}. These experiments will help us understand how much predictive power is contained in food words alone versus the full text (or hashtags) of the tweets, which capture a much broader context.

\subsection{Topic Model Features}
\label{sec:topical_features}

Topic models provide a method to infer the themes present in tweets, represented as clusters of words that tend to appear in similar contexts (e.g., a topic learned by the model, which we refer to as the \emph{American diet} topic, contains {\em chicken}, {\em baked}, {\em beans}, and {\em fried}, among other terms).
Using topics as features is beneficial for a couple of reasons:
(1)~topics provide a method to address the sparsity resulting from having very short documents (tweets are limited to 140 characters) by treating groups of related words as a single feature;
(2)~topical features aid in post-hoc analysis by allowing us to detect correlations that go beyond individual words.

We use Latent Dirichlet Allocation (LDA)\cite{blei2003latent} to learn a set of topics from the food tweets in an unsupervised fashion. 
LDA treats each tweet as a mixture of latent topics. 
Each topic is itself a probability distribution over words, and the words in the tweet are viewed as being sampled from this mixture of distributions.
The LDA model (topic distributions and mixtures) is trained from all available tweets in the corpus, using the \textsc{Mallet} software package.\footnote{\url{http://mallet.cs.umass.edu/}} 
We chose 200 as the number of topics for the model to learn.
This number produced topics that seemed fine-grained enough to capture specific patterns in diet, language, or lifestyle -- clusters of foods of various nationalities, or specific diets such as vegetarian.
For clarity in our analysis, we have manually assigned subject labels, such as {\em American diet}, to some of these topics based on the words contained in the topic.\footnote{But these topic labels are not visible to the classifier.}
We use these assigned labels to refer to the topics in the remainder of this paper.
Once the LDA topic model is trained, we use it to infer the mixture of topics for each tweet in the prediction tasks. The topic most strongly associated with the tweet (the topic with highest probability given the model and the tweet) is used as an additional feature for the tweet, similarly to the lexical features generated from the words of the tweet. Topics are counted across all tweets in a state in the same manner as the lexical features.

When applied in combination with the configuration containing solely food word or hashtag vocabularies, the LDA topics are constructed using the corresponding filtered versions of the tweets, i.e., with all non-food words or non-hashtag words removed.

To account for the large differences in the number of tweets available for each state (for example, the state with the most normalized tweets, New York has 83,670 tweets, while the state with the fewest, Wyoming, has 339), we scale all the features collected for each state. Each feature's value within a state's feature set is divided by the number of tweets collected for the state. 

\section{Results}

We present empirical results for both categories of tasks introduced
in the previous section: predicting state-level characteristics and predicting locales. 
We also analyze the effectiveness of the language of food for these prediction tasks by examining the most important textual features in the classification models, and investigating the importance of open versus closed vocabularies.

\subsection{State-Level Characteristics}
\begin{table}[t!]
    \centering
    \small
    \begin{tabularx}{\linewidth}{Xllll}
        \hline
        & {\small overweight}   & {\small diabetes}     & {\small political}    & {\small average}\\
        \hline
        \hline
        majority baseline     & 50.98        & 50.98        & 50.98        & 50.98 \\
        \hline
        All Words             & 76.47\sig  & 64.71  & 66.67\sig  & 69.28\sig  \\
        All Words + LDA       & {\bf 80.39}\sig  & 64.71  & 68.63\sig  & {\bf 71.24}\sig  \\
        \hline
        Hashtags              & 72.55\sig  & {\bf 68.63}\nearsig  & 60.78  & 67.32\sig  \\
        Hashtags + LDA        & 74.51\sig  & {\bf 68.63}\nearsig  & 62.75  & 68.63\sig  \\
        \hline
        Food                  & 70.59\sig & 60.78  & 68.63\sig & 66.67\sig \\
        Food + LDA            & 68.63\nearsig & 60.78  & {\bf 72.55}\sig & 67.32\sig \\
        \hline
        Food + Hashtags       & 64.71\nearsig & 62.75  & 64.71\nearsig & 64.05\sig \\
        Food + Hashtags + LDA & 74.51\sig\ssig & 62.75  & 64.71\nearsig & 67.32\sig\nnearsig \\
        \hline
    \end{tabularx}
\caption{\small \label{tbl:dataset_results} {\small
    Using features of tweets to predict state-level
    characteristics: whether a given state is above or below the
    national median for overweight rate, above or below the median
    diagnosed diabetes rate, and the state's historical political
    voting trend (D or R).
    This table compares the effect of filtering the lexical features
    to: food words, hashtags, both, or keeping the entire text of the tweets;
    as well as the effect of adding LDA topics.
Throughout the paper, we mark results 
as follows: 
\sig denotes a significant $(p <= 0.05)$  and
\nearsig a nearly-significant $(0.05 < p <= 0.10)$ improvements over the majority
baseline.
Similarly, \ssig denotes that the LDA model has a statistically
significant $(p <= 0.05)$ and \nnearsig a nearly statistically significant $(0.05 < p <= 0.10)$ improvement over the model without LDA. 
Statistical significance testing is implemented using one-tailed, non-parametric bootstrap resampling with 10,000 iterations.
}
}
\end{table}
Table~\ref{tbl:dataset_results} shows classification results on the state-level statistics prediction task (Section~\ref{sec:state_level}) for varying feature sets.
Since all three datasets are nearly evenly split between the binary classes (each dataset has either 25 or 26 states out of 51 in each of the two classes), a baseline that predicts the majority label achieves approximately 51\% accuracy.
We compare the performance of the tweet-based predictive models to this majority baseline, and evaluate how filtering the lexical content of the tweets and adding topical features affects accuracy on these prediction tasks.
\begin{table}[bt]
    \centering
    \begin{small}
    \begin{tabular}{lp{60mm}} %
        \hline
        Class & Highest-weighted features \\
            \hline
            \hline
            overweight: +& i, day, my, great, one, \emph{American Diet} (chicken, baked, beans, fried), \#snack, \emph{First-Person Casual} (my, i, lol), cafe, \emph{Delicious} (foodporn, yummy, yum), \emph{After Work} (time, home, after, work), house, chicken, fried, {\emph Breakfast} (day, start, off, right), \#drinks, bacon, call, eggs, broccoli \\
            \hline
            overweight: -& \emph{You, We} (you, we, your, us), \#rvadine, \#vegan, make, photo, dinner, \#meal, \#pizza, \emph{Giveaway} (win, competition, enter), new, \emph{Restaurant Advertising} (open, today, come, join), \#date, happy, \#dinner, 10, jerk, check, \#food, \#bento, \#beer \\
            \hline
            \hline
            diabetes: +& \emph{Mexican} (mexican, tacos, burrito), \emph{American Diet} (chicken, baked, beans, fried), \#food, \emph{After Work} (time, home, after, work), \#pdx, my, lol, \#fresh, \emph{Delicious} (foodporn, yummy, yum), \#fun, morning, special, good, cafe, \#nola, fried, bacon, \#cooking, all, beans \\
            diabetes: -& \#dessert, \emph{Turkish} (turkish, kebab, istanbul), \#foodporn, \#paleo, \#meal, \emph{Paleo Diet} (paleo, chicken, healthy), i, \emph{Giveaway} (win, competition, enter), \emph{I, You} (i, my, you, your), your, new, today, \#restaurant, \emph{Japanese} (ramen, japanese, noodles), some, jerk, \#tapas, more, \emph{Healthy DIY} (salad, chicken, recipe), \emph{You, We} (you, we, your, us) \\
            \hline
            \hline
            Democrat & \#vegan, \#yum, w, served, \#brunch, \emph{Deli} (cheese, sandwich, soup), photo, \#rvadine, \emph{Restaurant Advertising} (open, today, come, join), \#breakfast, \#bacon, delicious, \#food, \#dinner, 21dayfix, like, \#ad, \emph{Giveaway} (win, competition, enter), toast, 1 \\
            Republican & my, \#lunch, i, \emph{Airport} (airport, lounge, waiting), easy, \#meal, tonight, \#healthy, \#easy, us, sunday, \emph{After Work} (time, home, after, work), \#party, \#twye, \emph{First-Person Casual} (my, i, lol), your, \#snack, join, \#delicious, house \\
            \hline
        \end{tabular}
        \end{small}
    \caption{\small \label{tbl:dataset_feature_weights}{\small Top 20 highest-weighted features in descending order of importance for each dataset, for both the positive and negative classes. For example, ``overweight: +'' indicates the most representative features for being overweight, whereas ``overweight: -'' shows the most indicative features for {\bf not} being overweight.
The features include LDA topics, with manually assigned names (\emph{italicized}) for clarity, and a few of their most common words within parentheses.}}
\end{table}
We draw several observations from this experiment:

\noindent (a) First and foremost, the language of food can indeed infer all the latent characteristics investigated: all configurations investigated statistically outperform the majority-class baseline. The best performance is obtained when the entire text of the tweets is used (All Words), which captures not only direct references to food, but also the context in which it is discussed. However, the performance of the closed vocabulary of food words (Food) is within 5\% of the best performance, demonstrating that most of the predictive signal is captured by direct references to food.

\noindent (b) The classifiers achieve the highest accuracy on the overweight dataset. This is an intuitive result, which confirms that there is a strong correlation between food and likelihood of obesity. However, the fact that this correlation can be detected solely from social media posts is, to our knowledge, novel and suggests potential avenues for better and personalized public health. 
A similar correlation with political preferences is also interesting, 
indicating potential marketing applications in the political domain.

\noindent (c) More complex NLP (topic modeling in our case) is beneficial: the performance of the models that include LDA topics is, on average, better than that of the configurations without topics.\footnote{The improvement is not statistically significant for most experiments, but this can be attributed to the small size of the dataset (51 data points).} We plan to use more informative representations of text, e.g., based on deep learning~\cite{socher2013}, in future work.

Table~\ref{tbl:dataset_feature_weights} shows the words and topical features assigned the greatest importance, i.e., largest magnitude weights, by the SVM training process, for each dataset and class.
It is interesting to note that a dietary topic we have labeled as
\emph{American Diet}, containing terms such as {\em chicken}, {\em baked}, {\em
beans} and {\em fried}, is an important feature for predicting both
that a state has higher rates of overweight and diabetes than normal,
whereas other diets, such as {\em \#vegan} and {\em Paleo Diet}
are important predictors for the opposite. 
Note that pronouns have high weights in the overweight prediction task: the first-person singular {\em I} and {\em my} are valuable for predicting that a state is overweight, while collective words such as the \emph{You, We} topic cluster are valuable for predicting that a state is below the median.
This is less surprising in view of prior work, such as Ranganath et al.~\cite{ranganath2009s}, showing that the types of pronouns used by an individual are associated with a host of traits such as gender and intention.
For the political affiliation task, we observe that features
correlated with Republican states include those centered around work
(the \emph{Airport} topic) and home (the \emph{After Work} topic, including words such as {\em home}, {\em after}, {\em work}).
The most predictive feature for Democratic states is {\em \#vegan}, and we also see topics associated with urban life and eating out, such as \emph{Deli}, {\em \#brunch}, promotions such as \emph{Restaurant Advertising}, and \emph{Eating Out}. 

\subsection{City Prediction}
\begin{table}
    \centering
    \small
    \begin{tabular}{ll}
        \hline
        model & accuracy (\%) \\
    \hline \hline
        Random Baseline          & 6.67 \\
        \hline
        All Words         & 66.67\sig \\
        All Words + LDA     & 80.00\sig\nnearsig \\
        \hline
        Food              & 40.00\sig \\
        Food + LDA          & 40.00\sig \\
        \hline
        Hashtags          & 53.33\sig \\
        Hashtags + LDA      & 66.67\sig \\
        \hline
        Food + Hashtags     & 53.33\sig \\
        Food + Hashtags + LDA & {\bf 86.67\sig\ssig} \\
        \hline
    \end{tabular}
    \caption{\small \label{tbl:city_results}{\small City prediction accuracy (15 most populous US cities) for the various feature sets.
    Statistical significance testing is performed similarly to Table~\ref{tbl:dataset_results}.
}}
\end{table}
For the first locale prediction task we focus on city identification.
Table~\ref{tbl:city_results} shows the accuracies of the various feature sets for this task.
The input for this task is 15 cities, so the random-prediction
baseline accuracy is 6.67\%.
As in the previous task, every set of features improves significantly upon this baseline, ranging from 40\% accuracy using only Food words to 86.67\% accuracy using Food words, Hashtags, and LDA topics, demonstrating once again the predictive power of the language of food.
The significant improvement of the closed food vocabulary alone (Food) over the baseline indicates that the diets in each of these 15 cities are distinct enough to have some predictive power.
However, diets alone are not enough to completely identify the cities, and we see that for this task more context is beneficial: adding hashtags helps considerably (53.33\% accuracy), and adding topical features to the food and hashtag filtered set of lexical features improves performance even further (86.67\%). 
\begin{table}
\small
\centering
\begin{tabular}{r|rrrrr}
    \hline
    & \multicolumn{5}{c}{testing fraction} \\
 training fraction & 0.2 & 0.4 & 0.6 & 0.8 & 1.0 \\
    \hline
    0.2 & 6.66 & 6.66 & 6.66 & 6.66 & 6.66 \\
    0.4 & 13.33 & 13.33 & 13.33 & 13.33 & 20.00 \\
    0.6 & 20.00 & 26.66 & 26.66 & 26.66 & 40.00 \\
    0.8 & 33.33 & 46.66 & 33.33 & 53.33 & 53.33 \\
    1.0 & 46.66 & 53.33 & 60.00 & 66.66 & 80.00 \\
    \hline
\end{tabular}
\caption{\small \label{tbl:city_learning_curves}Effects of varying the fraction of tweets used for training and testing on classification accuracy in the city-prediction task, using All Words and LDA topics.}
\end{table}
\begin{table}[t]
\small
    \begin{tabularx}{\linewidth}{lX}
        \hline
        City & Highest-weighted features \\
        \hline
        Austin &  we, come, tacos, \#tacos, \emph{Mixed Drinks} (bottomless mimosas, bloody mary) \\
        Chicago &  \emph{Giveaway} (win, competition, enter), jerk, \#breakfast, \#bbq, \#foodie \\
        Columbus &  \#breakfast, \#asseenincolumbus, \emph{Directions} (west, local, east), \#cbus, \#great \\
        Dallas &  \#lunch, my, lunch, porch, come \\
        Houston &  \emph{After Work} (time, home, after, work), \#lunch, \#snack, i, \#breakfast \\
        Indianapolis &  you, our, delicious, \emph{You, We} (you, we, your, us), side \\
        Jacksonville &  \#dinner, \#ebaymobile, \#food, kitchen, \#yum \\
        Los Angeles &  my, \#foodie, \emph{Directions} (west, local, east), \#timmynolans, \#tolucalake \\
        New York City &  \#brunch, \emph{Mixed Drinks} (bottomless mimosas, bloody mary), our, \emph{Eggs and Bacon} (eggs, benedict, bacon), \#sarabeths \\
        Philadelphia &  cafe, day, \#fishtown, shot, \#byob \\
        Phoenix &  \#lunch, \#easy, \emph{Wine} (wine, 2014, today), st, we \\
        San Antonio &  my, i, 1, bottomless, our \\
        San Diego &  \emph{Restaurant Advertising} (open, today, come, join), \#bottomless, \emph{Mixed Drinks} (bottomless mimosas, bloody mary), \emph{Vacation} (beach, hotel, vacation), your \\
        San Francisco &  \#vegetarian, \#dinner, \#foodie, brunch, \emph{Vacation} (beach, hotel, vacation) \\
        San Jose &  \#foodporn, \#dinner, bill, \#bacon, \#goodeats \\
        \hline
    \end{tabularx}
        \caption{\small \label{tab:top5cities}{\small Top five highest-weighted features for predicting each city from its tweets. Features include All Words and LDA topics. LDA topics are manually assigned names (\emph{italicized}) for clarity, and a few of the most common words are displayed next to each topic.}}
 \end{table}

Table~\ref{tbl:city_learning_curves}, which measures performance as the size of training/testing sets varies, indicates that accuracy is greatly affected by the size of the data available for both training and testing.
Indeed, when only 20\% of the training set is used, the models achieve the same score as the baseline classifier (6.67\%). 
Performance continues to increase as we add more data, suggesting that we have not reached a performance ceiling yet.

Table~\ref{tab:top5cities} lists the top five features for each city in this task.
The table shows that variations in diet are clear: {\em tacos} are significant in Austin, {\em \#vegetarian} food is indicative of San Francisco, {\em \#brunch} is representative of New York, etc. 
Using the context around food is clearly important.
We see that several cities in California are associated with {\em \#foodie} (Los Angeles and San Francisco) or  eating while on \emph{Vacation} (San Diego and San Francisco). 
First-person pronouns are highly weighted in cities in Texas ({\em we} in Austin, {\em I} in Houston, and {\em my} and {\em I} in San Antonio). 
\subsection{State Prediction}

\begin{table}[t!]
    \centering
    \begin{small}
\begin{tabular}{ll}
    \hline
    model                 & accuracy (\%) \\
    \hline \hline
     Random Baseline      & 1.96 \\
    \hline
    All Words             & 60.78\sig \\
    All Words + LDA       & {\bf 66.67}\sig\ssig \\
    \hline
    Food                  & 33.33\sig \\
    Food + LDA            & 35.29\sig \\
    \hline
    Hashtags              & 62.75\sig \\
    Hashtags + LDA        & 56.86\sig \\
    \hline
    Food + Hashtags       & 56.86\sig \\
    Food + Hashtags + LDA & 54.90\sig \\
    \hline
\end{tabular}
\end{small}
\caption{\small \label{tbl:state_results}{\small State prediction accuracy for the various features sets.
}}
\end{table}
Table~\ref{tbl:state_results} lists the results for the state prediction task. There are 51 possible locales in this task, from the 50 US states plus Washington D.C., so the random-prediction baseline achieves 1.96\% accuracy. As in all previous experiments, the state-prediction model improves significantly upon this baseline with every set of features that we tried. The model achieves its lowest accuracy, 33.33\%, using the set of food words without topical features (Food). Unlike in the city prediction task but similar to the state-level characteristic prediction task, the model is most accurate when using the unfiltered tweets with topical features (All Words + LDA), reaching 66.67\% accuracy. This indicates that the closed food vocabulary is not sufficient for optimal performance on this task, and the larger food-related context is required for optimal performance. 
\begin{table}
\small
\begin{tabular}{r|rrrrr}
    \hline
    & \multicolumn{5}{c}{testing fraction} \\
 training fraction & 0.2 & 0.4 & 0.6 & 0.8 & 1.0 \\
    \hline
    0.2 & 11.76 & 11.76 & 5.88 & 9.80 & 15.68 \\
    0.4 & 19.60 & 17.64 & 17.64 & 17.64 & 25.49 \\
    0.6 & 25.49 & 29.41 & 35.29 & 41.17 & 47.05 \\
    0.8 & 39.21 & 41.17 & 43.13 & 50.98 & 52.94 \\
    1.0 & 43.13 & 58.82 & 54.90 & 62.74 & 64.70 \\
    \hline
\end{tabular}
    \caption{\small \label{tbl:state_learning_curves}{\small Effects of varying the fraction of tweets used for training and testing on classification accuracy in the state-prediction task, using All Words and LDA topics.}}
\end{table}

Table~\ref{tbl:state_learning_curves} analyzes the effect of the number of available tweets on prediction accuracy.
Performance varies from 11.76\% when using 20\% of tweets in the training set and 20\% in the testing set, to 64.7\% when using all available tweets.
Increasing the number of tweets in the training set has a larger positive effect on accuracy than increasing the number of tweets in the testing set.
As in the city prediction task, performance continues to increase as we add more data, suggesting that the performance ceiling has not been reached.

Due to space consideration, we include the top features per state and
the corresponding discussion in the supplemental material.\footnote{\url{https://sites.google.com/site/twitter4food/}}

\subsection{Region Prediction}

The final locale prediction task predicts the four major US geographic regions: Midwest, Northeast, South, and West (Section~\ref{sec:task_locale_region}) using tweets from each region.
The high level of geographic granularity (each region contains about a dozen states, on average) simplifies the tweet-based task in one sense, since there are now fewer possible classification labels, but also makes the task more difficult because of the variation in diet and tweet lexical content within these broad geographic regions. 
\begin{table}
\small
    \centering
    \begin{tabular}{ll}
        \hline
        model & accuracy (\%)  \\
        \hline \hline
        Random Baseline & 25 \\
        \hline
        All Words& 50  \\
        All Words + LDA & {\bf 75}  \\
        \hline
        Food & 50  \\
        Food + LDA & 50  \\
        \hline
        Hashtags & 50  \\
        Hashtags + LDA & {\bf 75}  \\
        \hline
        Food + Hashtags & 50  \\
        Food + Hashtags + LDA & {\bf 75}  \\
        \hline
    \end{tabular}
 \caption{\small \label{tbl:region_results}{\small Region prediction accuracy for the various feature sets.
}}
 \end{table}
The random-prediction baseline in this task achieves 25\% accuracy.
Three of the feature sets, however, achieve 75\% accuracy, only misclassifying a single region.\footnote{Since there are only four data points in the testing set, measuring if improvements over the baseline are statistically significant cannot be reliable, so it is skipped here.}
We also see that for all of the feature sets except the closed food vocabulary (Food), lexical features give one more correct region classification over the baseline, and adding topical features yields an additional correct classification. The feature set consisting of all words and the LDA topics classifies three out of four regions correctly, only misidentifying testing tweets from the Midwest as being from the Northeast. %

\begin{figure}
    \centering
        \includegraphics[width=.8\linewidth]{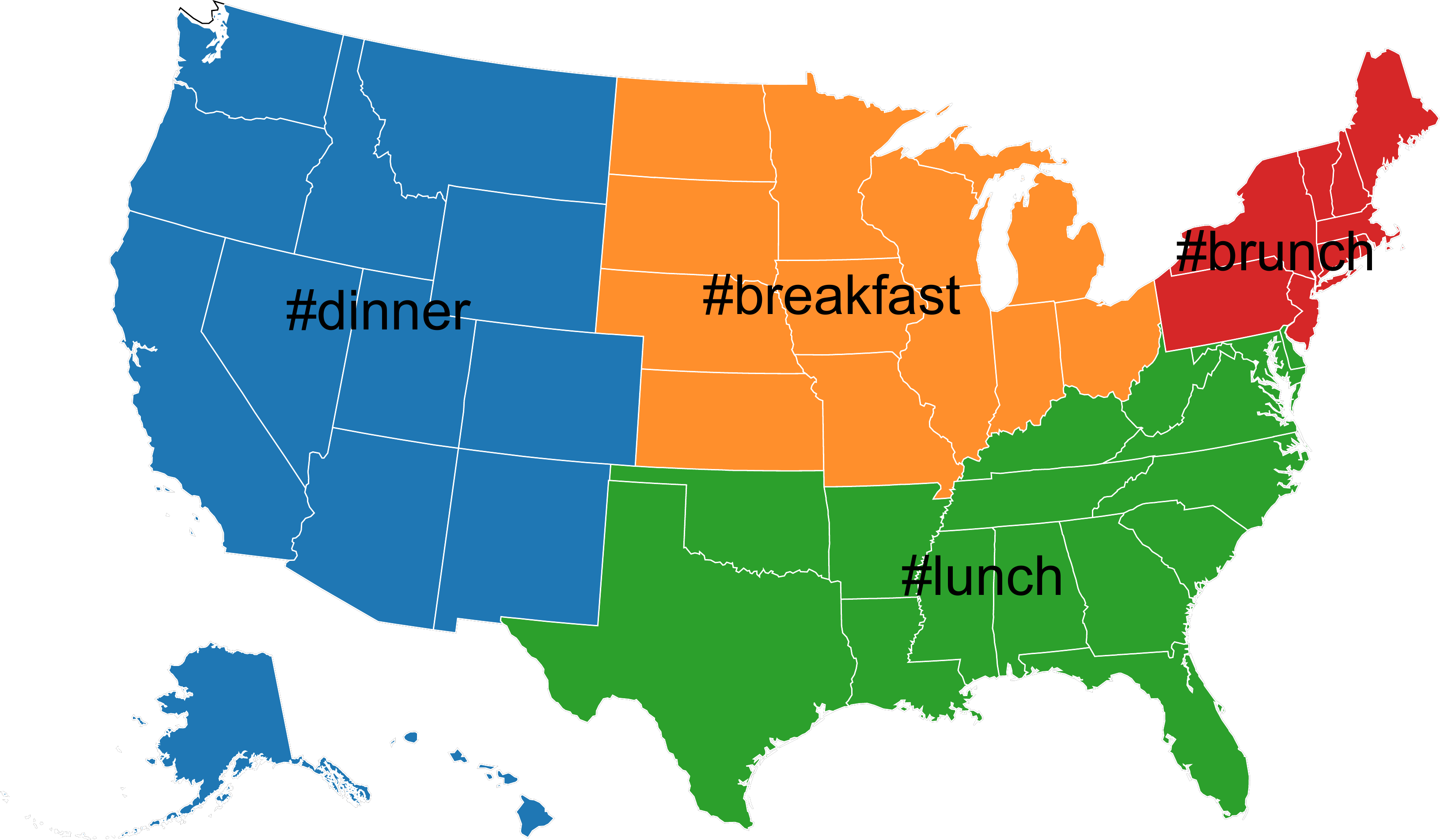}
        \caption{\small \label{fig:region_features}The highest-weighted feature for the region prediction task in each of the four US Census geographic regions: West (blue), Midwest (orange), South (green), and Northeast (red).}
\end{figure}

\begin{table}
\small
    \begin{tabularx}{\linewidth}{lX}
        \hline
        Region & Highest-weighted features \\
        \hline 
    Midwest &  \#breakfast, i, \#recipes, \emph{After Work} (time, home, after, work), \emph{Recipe} (recipe, easy, meal), your, \#meals, breakfast, \emph{Promotional} (free, off, today), \emph{I, You} (i, my, you, your)\\
    Northeast &  \#brunch, brunch, our, \emph{Mixed Drinks} (bottomless mimosas, bloody mary), we, w, \emph{Roasted Meats} (pork, chicken, special), \emph{Group Dining} (our, us, join), you, new\\
    South &  \#lunch, \emph{Mixed Drinks} (bottomless mimosas, bloody mary), \emph{After Work} (time, home, after, work), \emph{American Diet} (chicken, baked, beans, fried), chicken, \#cltfood, mimosas, bottomless, us, my\\
    West &  \#dinner, \#food, \#foodporn, photo, dinner, w, \#vegan, \emph{Mexican} (mexican, tacos, burrito), \#bomb, \#pdx\\
        \hline
    \end{tabularx}
    \caption{\small \label{tbl:region_features}{\small Top 10 highest-weighted features for predicting each region from its tweets.
    Features include All Words and LDA topics.
LDA topics are manually assigned names (\emph{italicized}) for clarity, and a few of the most common words are displayed next to each topic.
}}
\end{table}

Table~\ref{tbl:region_features} shows the most predictive features for each region.
We see a clear preference for certain meal term hashtags in the table:
{\em \#breakfast} is a strong predictor for the Midwest, {\em
  \#brunch} for the Northeast, {\em \#lunch} for the South, and {\em
  \#dinner} for the West; see Fig.~\ref{fig:region_features}.
Brunch and mixed drinks are important features in the Northeast, likely because they were also highly weighted features for New York City, and New York City produced many of the tweets in this region.
The West is a mixture of features that were important for California, such as {\em \#foodporn} and {\em \#vegan}, and the diet of other states in the region, such as the {\em Mexican} food topical feature.
The Northeast and South are similar in that both have the \emph{Mixed Drink} topic and a meat-related topic, but differ in other features, such as the Northeast's {\em \#brunch} and {\em Group Dining}, and the South's {\em \#lunch} and {\em After Work} topic.

\section{Visualization Tools}
\begin{figure}[b!]
    \centering
    \begin{subfigure}{0.23\textwidth}
        \centering
        \includegraphics[width=\linewidth]{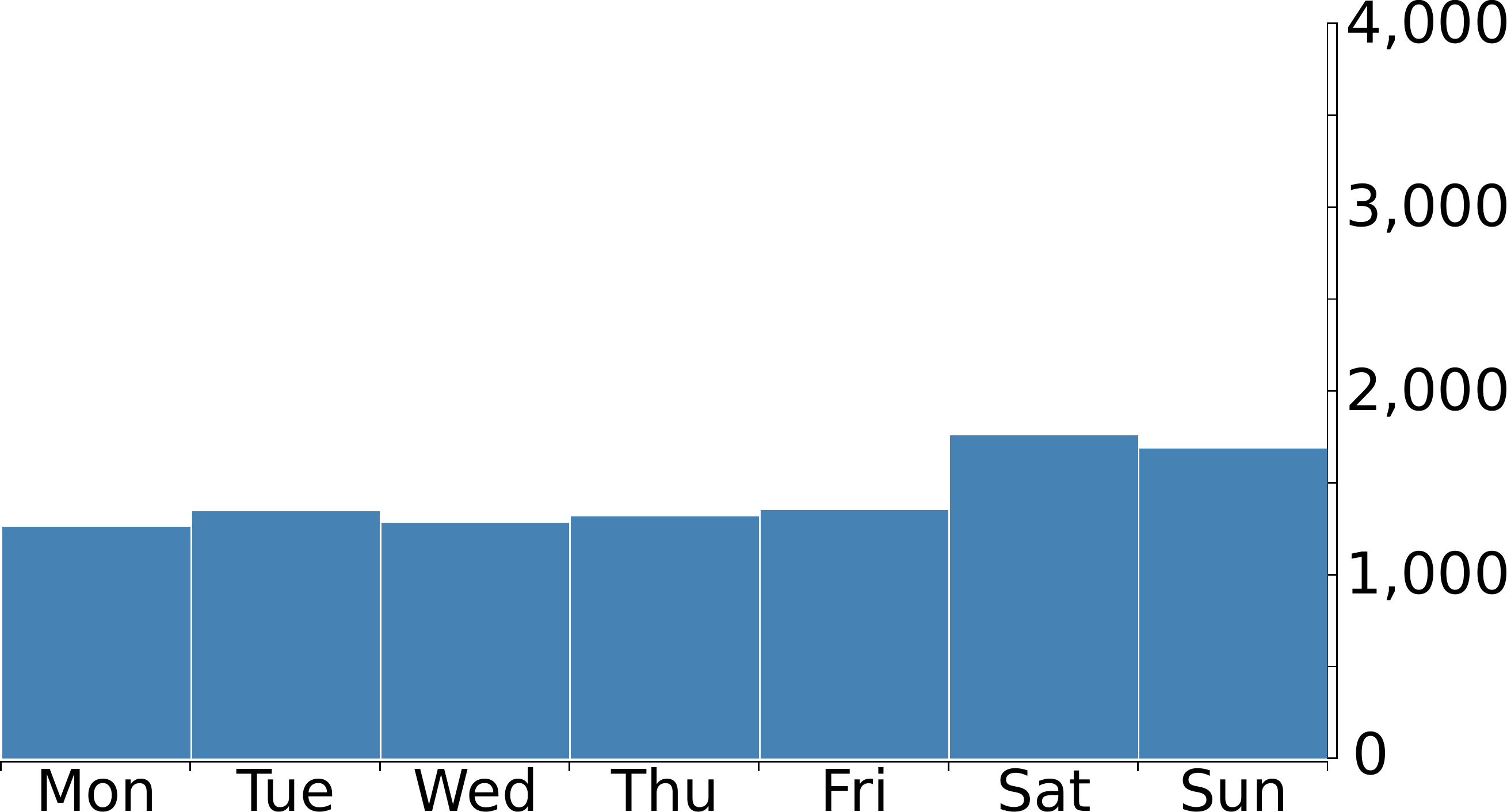}
        \caption{\small \label{fig:breakfast_hist}{\em breakfast} by day of post
}
    \end{subfigure}
    \begin{subfigure}{0.23\textwidth}
        \centering
        \includegraphics[width=\linewidth]{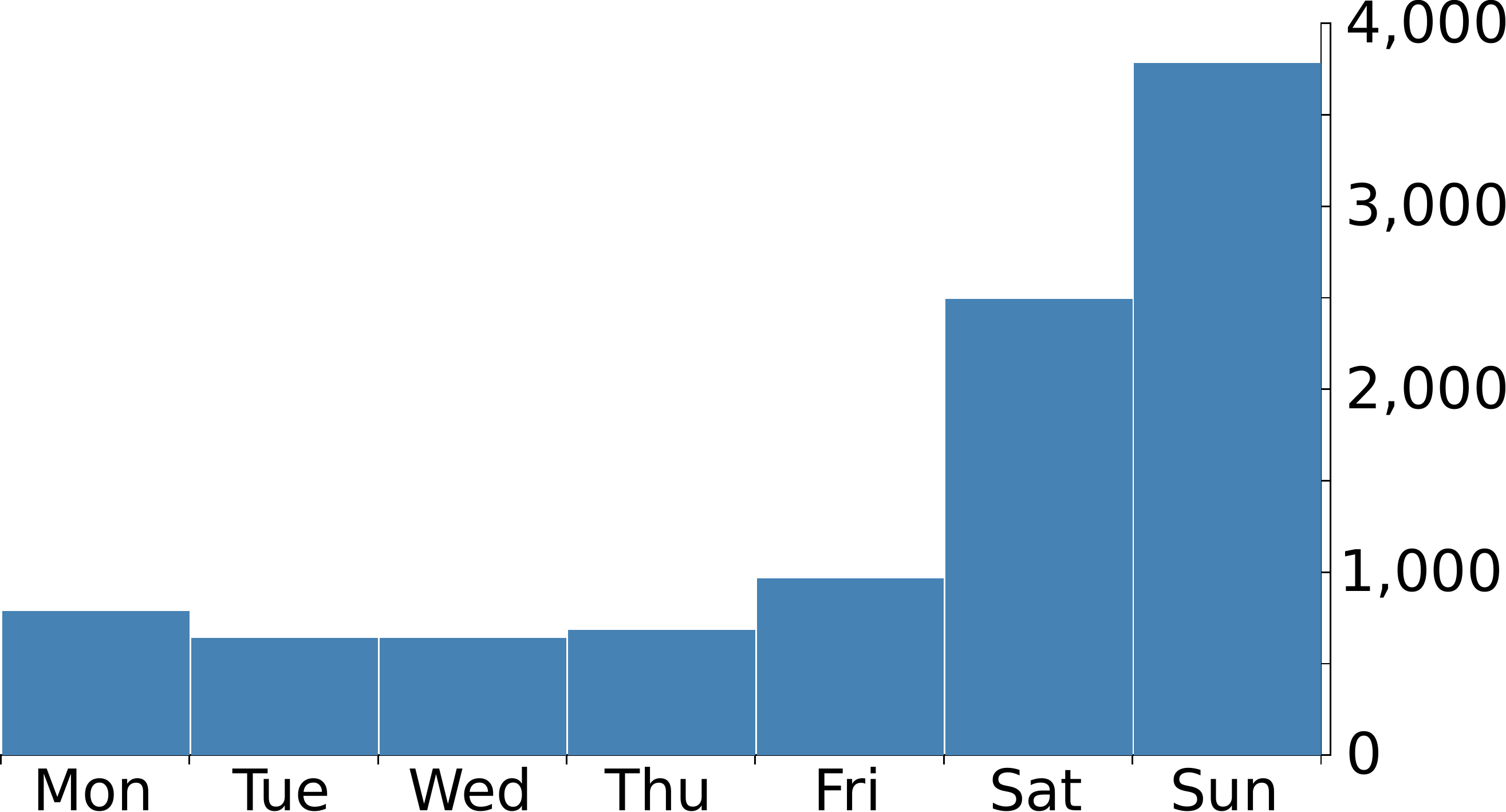}
        \caption{\small \label{fig:brunch_hist}{\em brunch} by day of post %
}
    \end{subfigure}

   \begin{subfigure}{0.23\textwidth}
        \includegraphics[width=\linewidth]{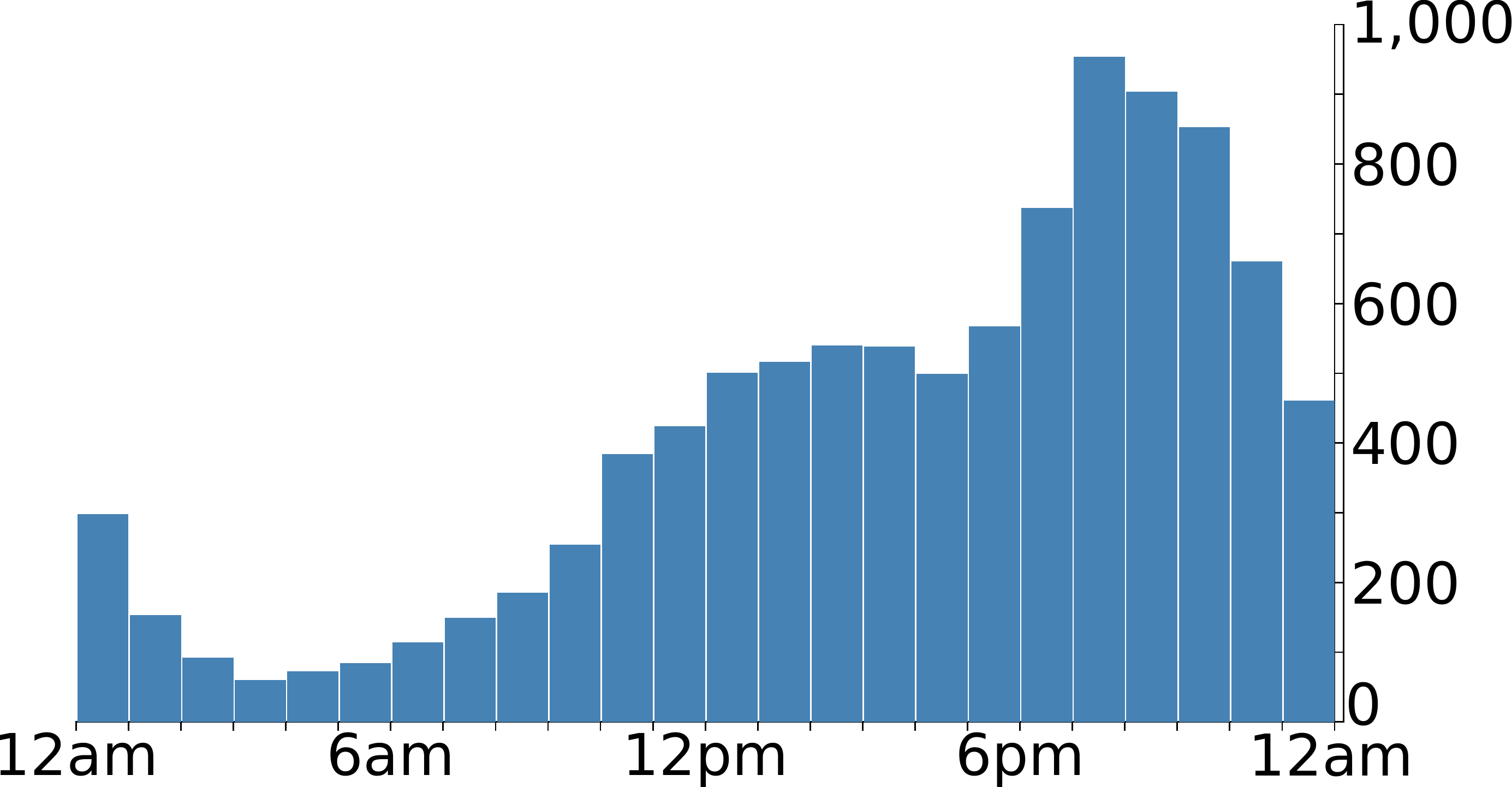}
        \caption{\small \label{fig:wine_hist}{\em wine} by hour of post}
    \end{subfigure}
    \begin{subfigure}{0.23\textwidth}
        \includegraphics[width=\linewidth]{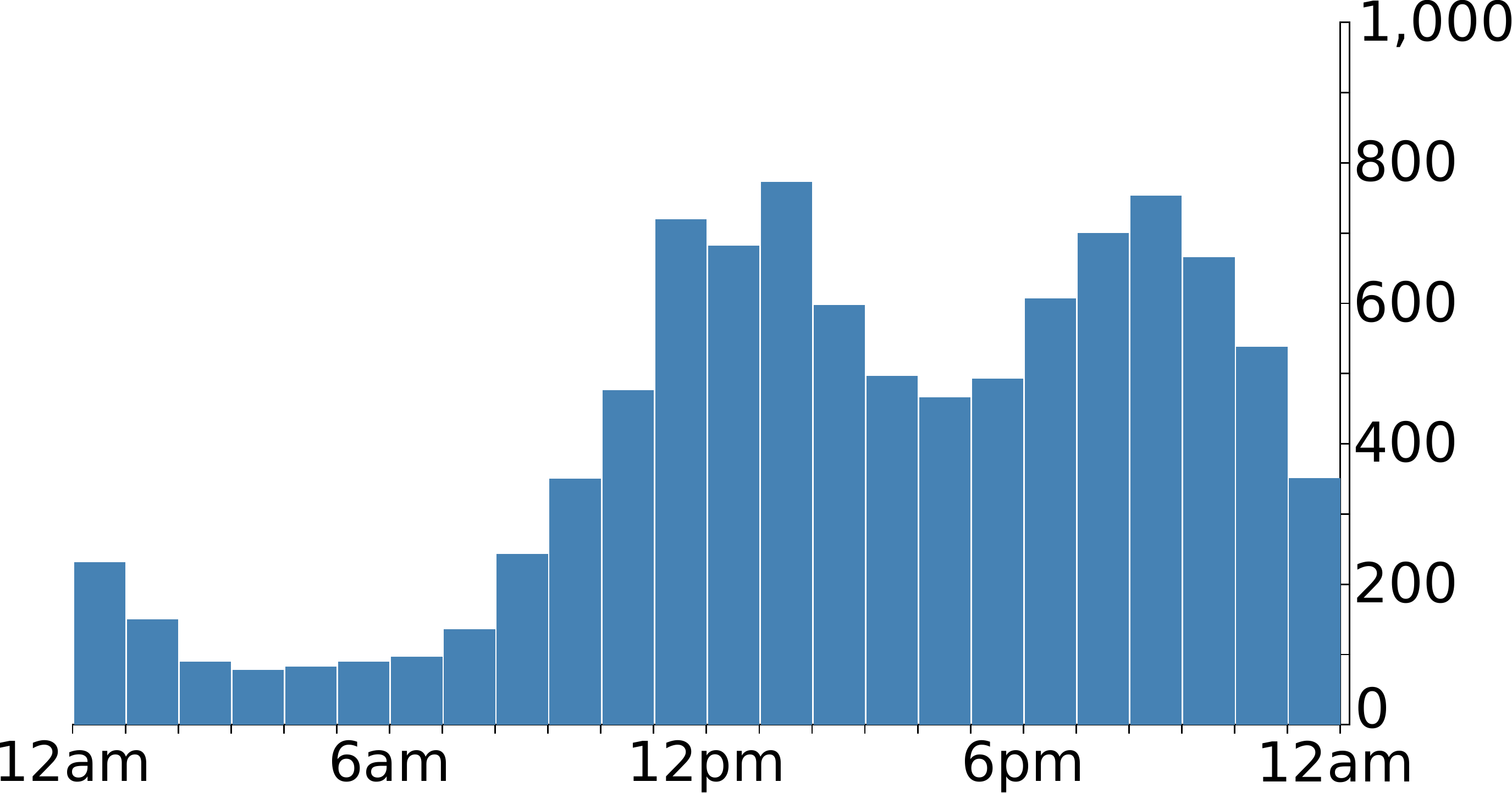}
        \caption{\small \label{fig:beer_hist}{\em beer} by hour of post}
    \end{subfigure}
    \caption{\small \label{fig:temporal_histograms}Temporal histograms showing the popularity of the phrases {\em breakfast} and {\em brunch} by day of the week, and {\em beer} and {\em wine} by hour of the day. For each term, 10,000 tweets are sampled from users who have listed their time zone, which is then used to obtain the local time of the tweet.}
\end{figure}

\label{viz}

While the machine learning models described above are well-suited for prediction on predefined tasks, we also constructed several visualization tools to discover previously unknown trends in the Twitter dataset.
These tools aim to allow aggregate analysis of tweet content in the context of geographic and temporal location.

\subsection{Top Terms by State}
The first of these tools, the term visualizer (Fig.~\ref{fig:terms_per_state}), does a simple keyword analysis of the tweets available for each US state.
We extract all terms that are contained within a list of around 800 food-related words
(see Sec.~\ref{sec:lexical_features}) and rank them using \emph{tf-idf}, treating all tweets normalized to a given state as a single document: each term's score is the number of times it occurred within a state, multiplied by the logarithm of the inverse proportion of the number of states it occurred in~\cite{manning2008iir}.
Ranking by \emph{tf-idf} emphasizes words that are common in a particular state, but ensures that words used frequently in all states, such as {\em food} and {\em eat}, are not highly ranked.
The term(s) with the highest ranking in each state are displayed on the state in the map.
As discussed previously, this tool immediately highlights dietary patterns: {\em grits} in the Southern states, etc.

\subsection{Temporal Histograms}
Temporal histograms allow us to visualize the changing popularity of terms over the course of a day, week, or year. 
About 71\% of the collected tweets (2,503,351) are from users who have listed their time zone. 
For these tweets, we compute the time local to the user when the tweet was posted. 
The temporal visualization tool (Fig.~\ref{fig:temporal_histograms}) allows querying these time-localized tweets by phrase and constructing histograms at varying time granularities: hour of day, day of the week, or month of the year. 
On the weekly scale, it is easy to see that while {\em breakfast} is more or less uniform all week long, {\em brunch} occurs much more frequently on weekends, particularly Sunday.
On the daily scale, {\em wine} peaks around 8pm, while {\em beer} follows a bi-modal distribution, with 
two roughly equal-sized peaks around 1pm and 8pm.

\subsection{Tweet Location Maps}
\begin{figure*}
    \centering
    \begin{subfigure}{\textwidth}
        \centering
        \includegraphics[width=0.45\linewidth]{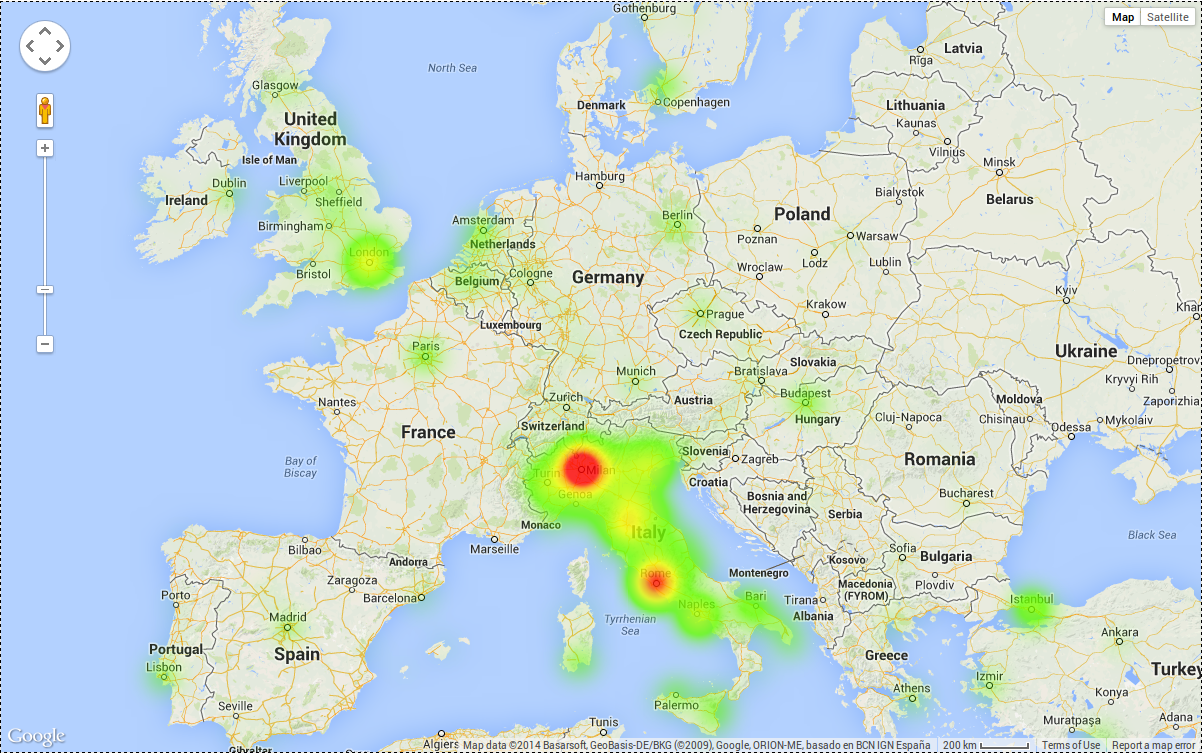}
        \hspace{6mm}
        \includegraphics[width=0.45\linewidth]{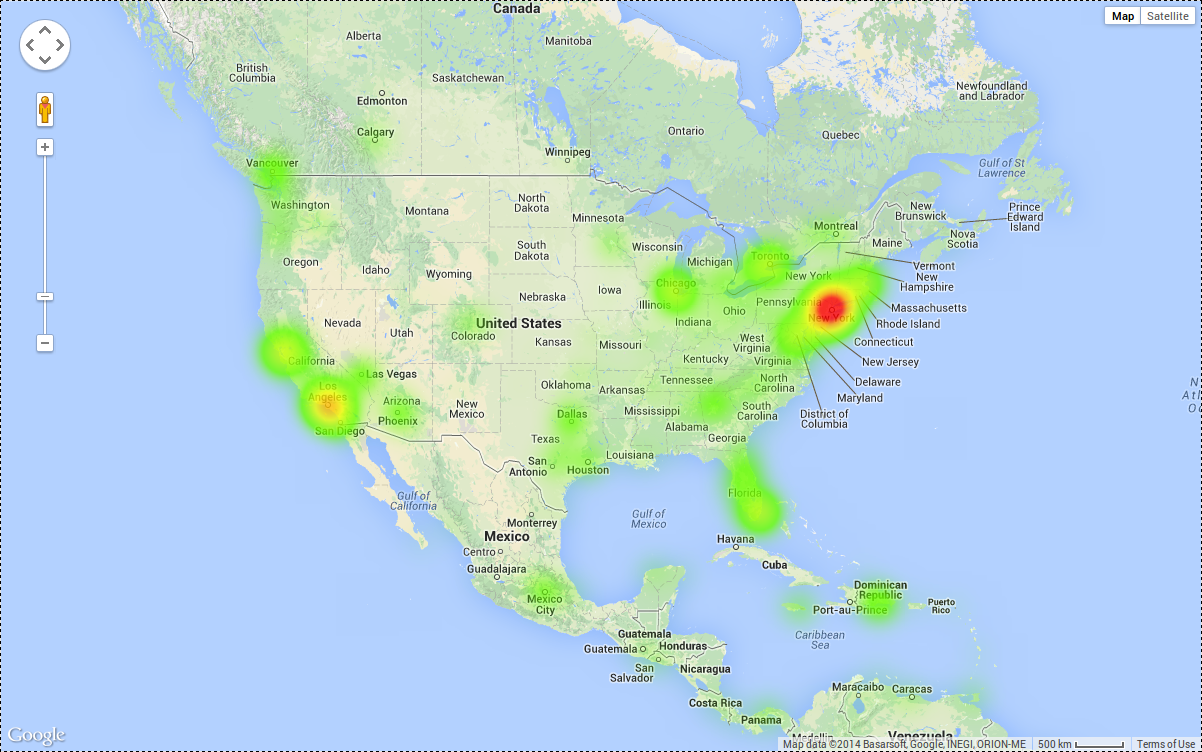}
        \caption{\small \label{fig:italian}Heatmaps of 7,372 tweets from three \emph{Italian food} ({\em pasta}, {\em pizza}, {\em italian}, {\em carbonara}, {\em lasagna}, ...) topics.} 
    \end{subfigure}
    \begin{subfigure}{\textwidth}
        \centering
        \includegraphics[width=0.45\linewidth]{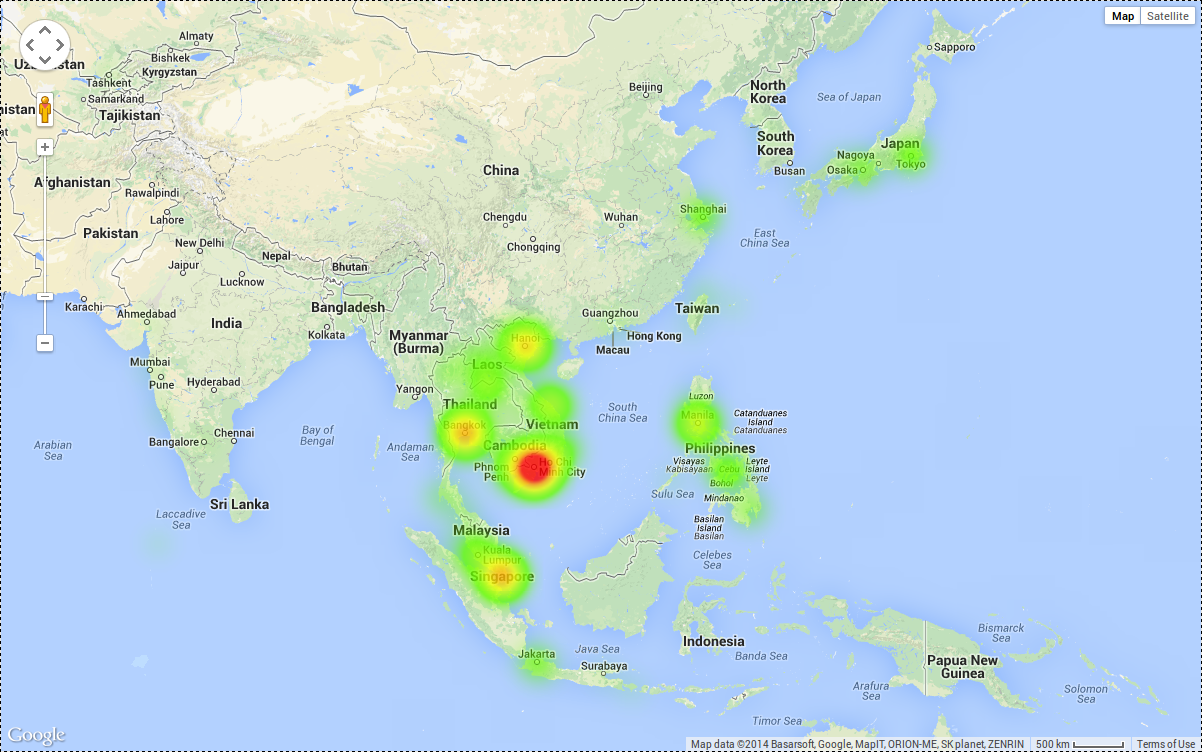}
        \hspace{6mm}
        \includegraphics[width=0.45\linewidth]{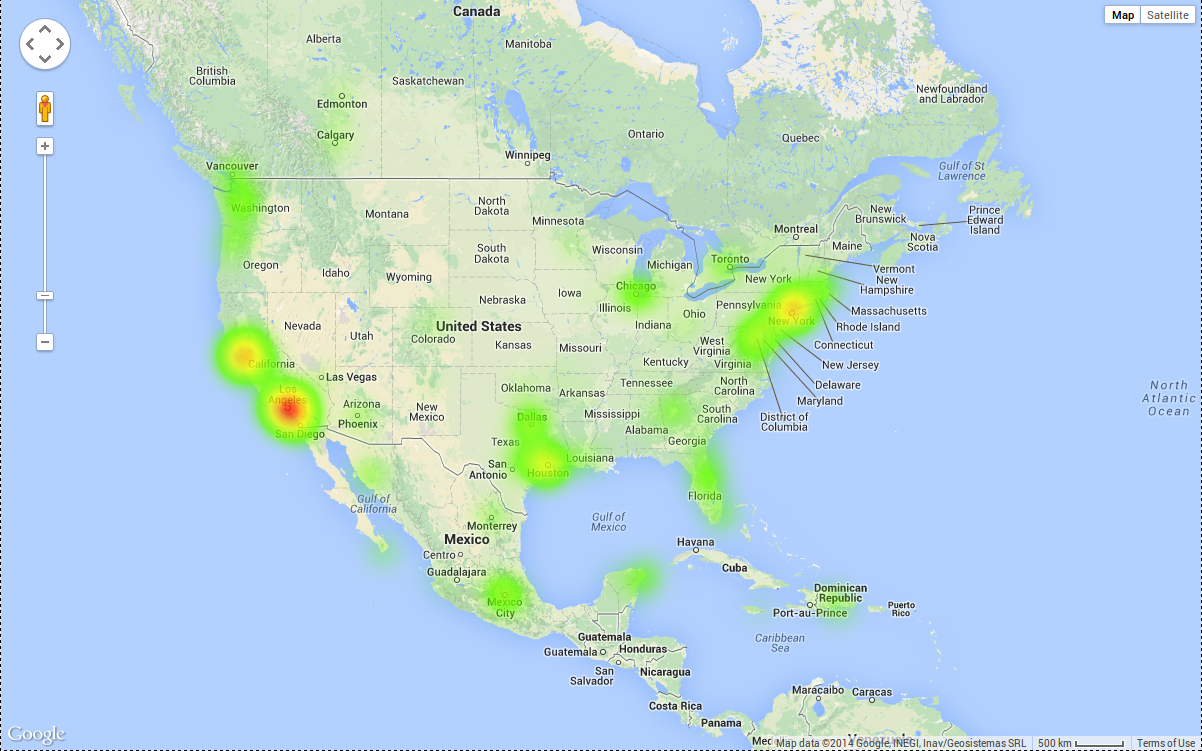}
        \caption{\small \label{fig:vietnamese}Heatmaps of 1,032 tweets from a \emph{Vietnamese food} ({\em pho}, {\em vietnamese}, ...) topic.}
    \end{subfigure}
   \begin{subfigure}{\textwidth}
        \centering
        \includegraphics[width=0.45\linewidth]{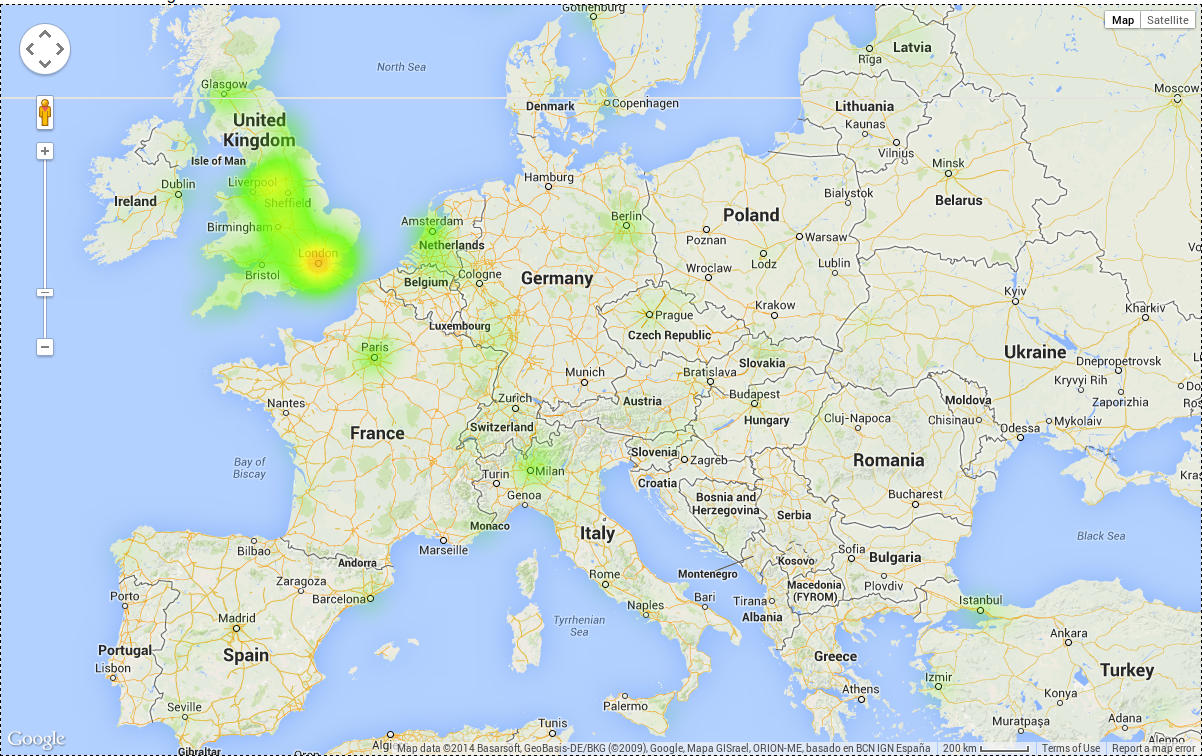}
        \hspace{6mm}
        \includegraphics[width=0.45\linewidth]{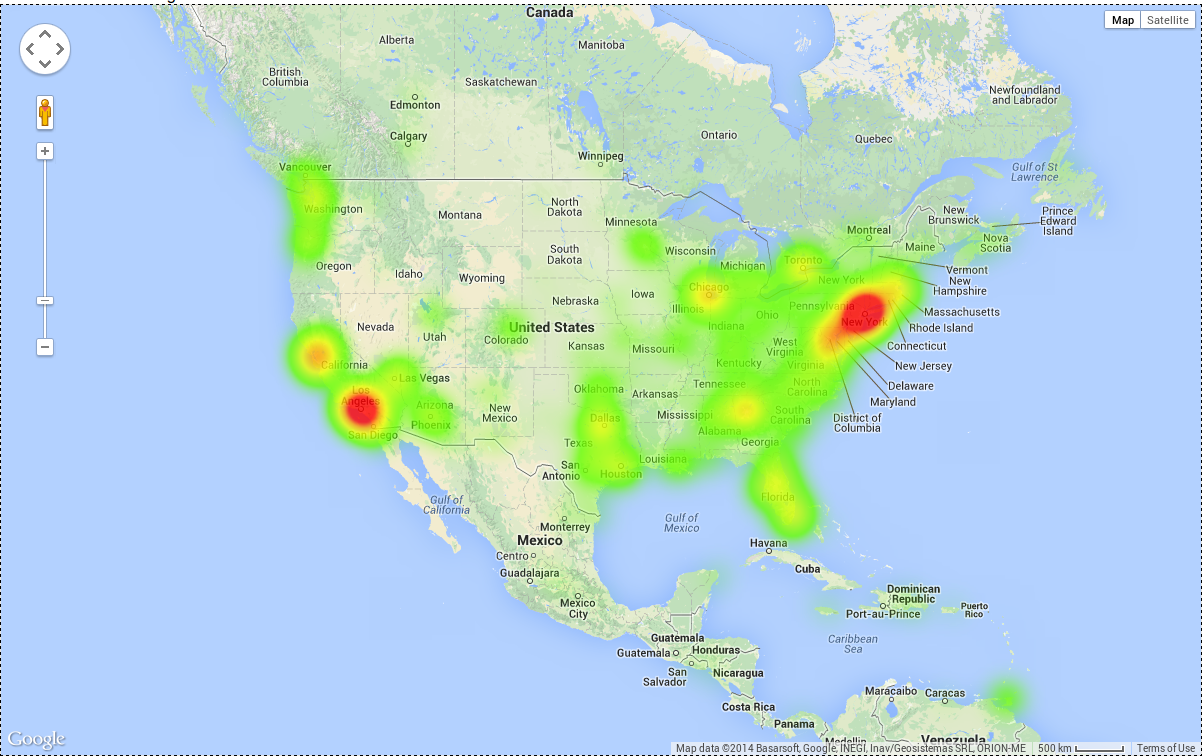}
        \caption{\label{fig:uk}Heatmaps of 2,226 tweets from a \emph{Full Breakfast} (``bacon'', ``eggs', ``sausage'', ``biscuits'', ``grits'', ...) topic.}
    \end{subfigure}
    \caption{\small \label{fig:migration1}Heatmaps showing migration patterns reflected in diet: Italian food has the largest concentration in the United States in New York City 
and Vietnamese food is highly concentrated in California, particularly Los Angeles and San Francisco.
}
\end{figure*}

\begin{figure}
        \includegraphics[width=\linewidth]{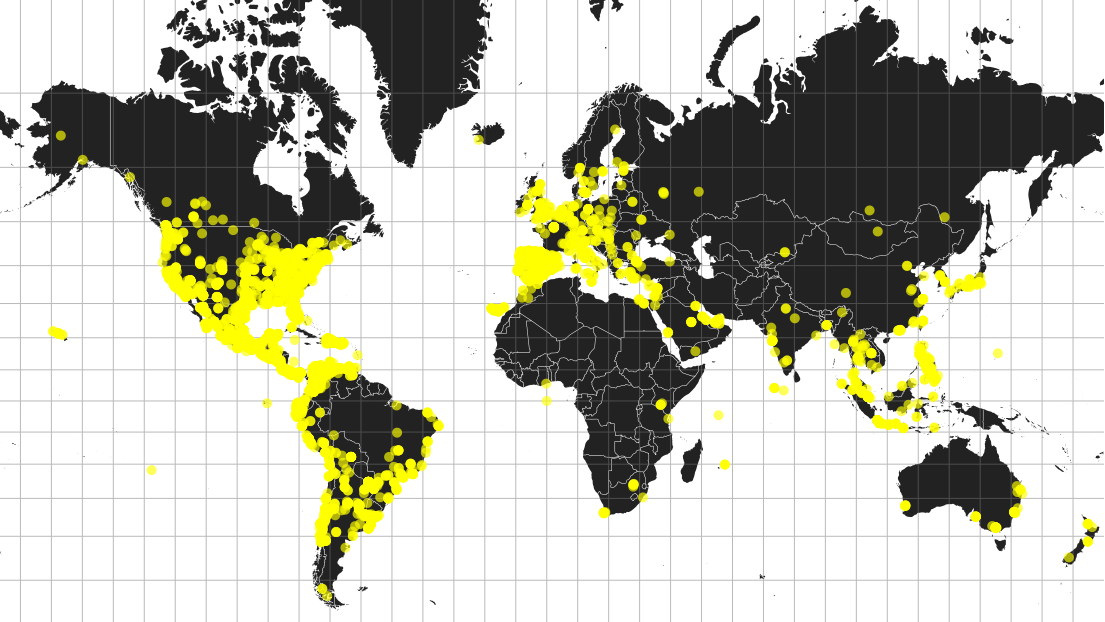}
    \caption{\small \label{fig:migration2}Tweet geolocation plot showing migration patterns reflected in diet: yellow dots mark the locations of 11,827 tweets matching five \emph{Spanish/Latin American food} topics ({\em tacos}, {\em burrito}, {\em salsa}, {\em pollo}, {\em arroz}, {\em paella}, etc.).}
    \vspace{-8mm}
\end{figure}

About 10\% of the collected tweets (362,978) have associated geolocation information -- the user's longitude and latitude at the moment the tweet was posted.
We use this meta-information to build a system for querying and plotting worldwide geographic maps of tweets.
The interface allows searching by phrase or LDA topic and displays geographic plots or heatmaps showing the locations of all tweets matching the query. 

This system allows the discovery of broad geographic trends in the data. For example, Fig.~\ref{fig:migration1} shows heatmaps made from queries for several LDA topics for foods of various geographic origins. 
These topics are perhaps reflective of immigration patterns to the US or worldwide, e.g., the \emph{Italian food} topic has high intensity in Italy and New York City (Fig.~\ref{fig:italian})  and the \emph{Vietnamese food} topic has high intensity in Vietnam and in Southern California (Fig.~\ref{fig:vietnamese}).
The \emph{Full Breakfast} topic, reflecting a traditional British and American breakfast of bacon, eggs, and sausage, is pronounced throughout all of English-speaking United Kingdom and the heavily populated regions of both the western and eastern United States (Fig.~\ref{fig:uk}). 
Similarly, Fig.~\ref{fig:migration2} shows the prevalence of Spanish and Latin-American influenced food throughout the Spanish-speaking world, including portions of the United States and the Philippines. 

\subsection{Parallel Word Clouds}
\begin{figure*}
\begin{center}
        \includegraphics[width=.4\linewidth]{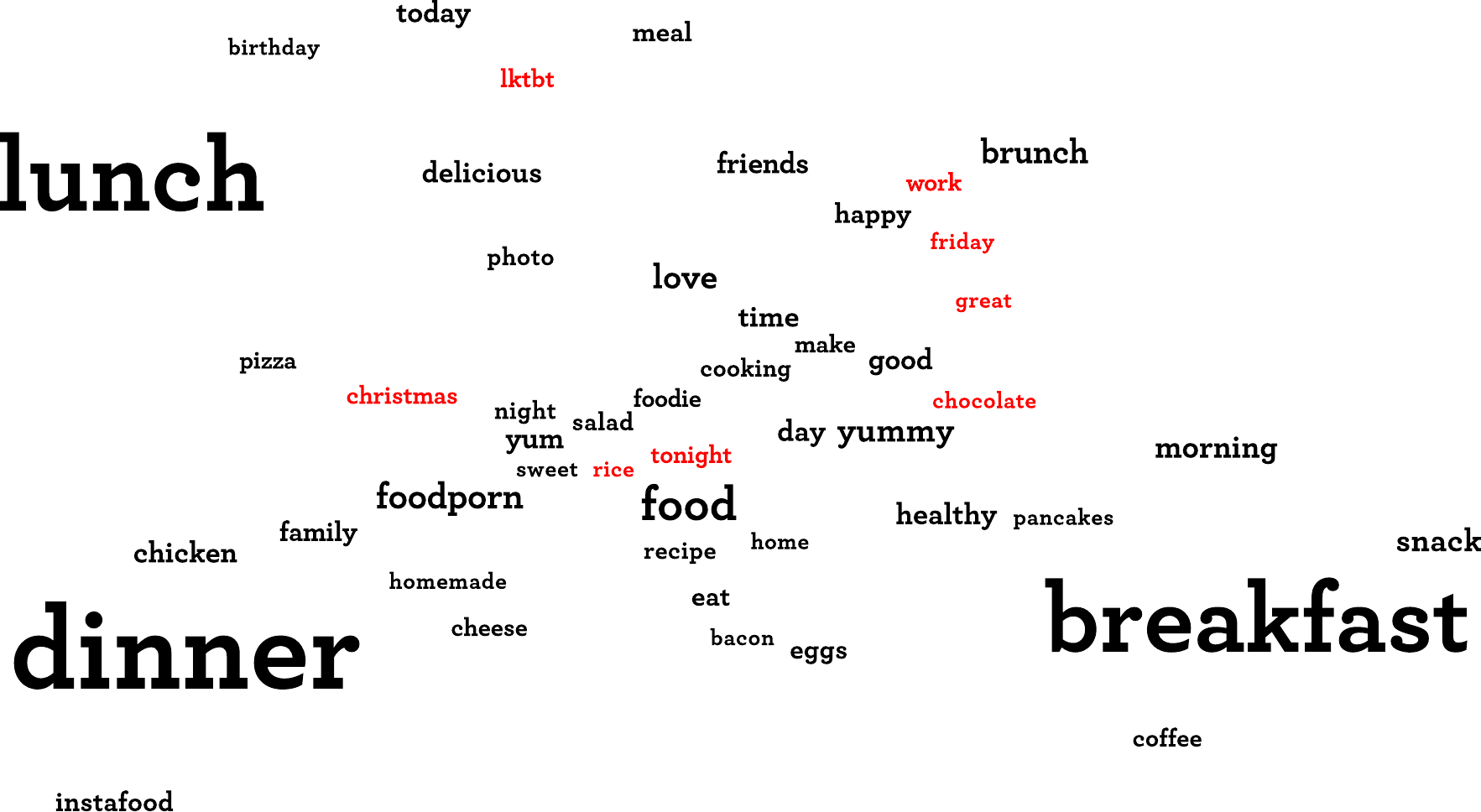}
    \hspace{0.4in}
        \includegraphics[width=.4\linewidth]{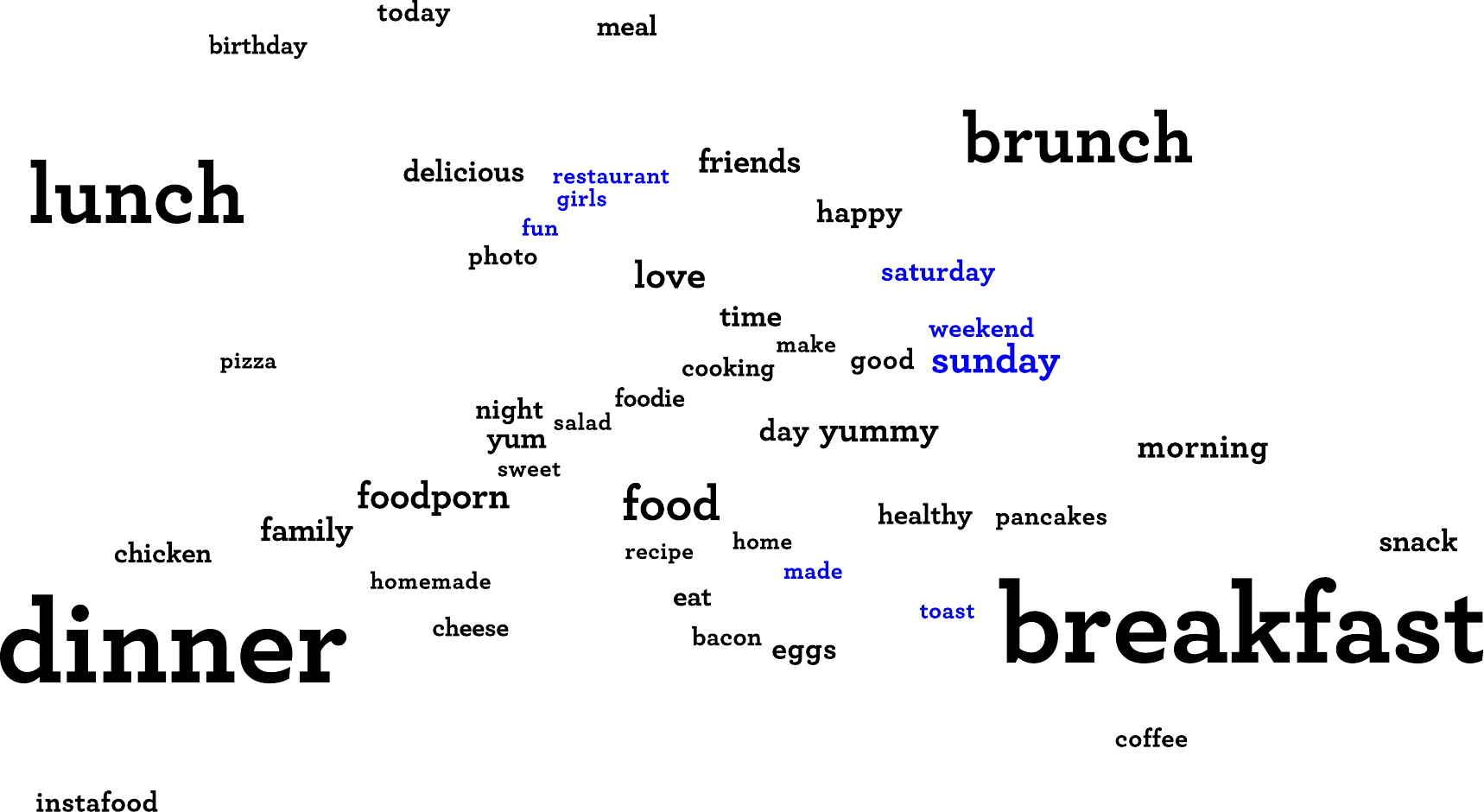}
\end{center}
    \caption{\small \label{fig:parallel_wc}Parallel semantics-preserving word clouds showing the difference between two sets of tweets: one containing weekday tweets (left) and one containing weekend tweets (right).
Words that appear frequently in both sets of tweets are black, those that appear frequently on weekdays are red, and those that are frequent on weekends are blue.}
\end{figure*}

Word clouds offer a space-efficient way to summarize text by highlighting important words. Semantics-preserving word clouds also add the feature that related words (e.g., those that frequently co-occur) are placed close to each other~\cite{barth14experimental}. Our parallel word clouds further focus on comparing and contrasting two or more groups of texts: words are scaled by importance, related words are close to each other, and important words that occur in both groups are in the same locations in all clouds ({\em citation hidden for review}). 
Fig.~\ref{fig:parallel_wc} shows such parallel word clouds for weekday vs. weekend tweets, highlighting a different set of trends:  {\em family},  {\em brunch}, and even {\em breakfast} are more prominent on weekends; {\em work} and {\em tonight} are common on weekdays (in red); and, finally, {\em restaurant}, and {\em fun} are present on weekends (in blue).

\section{Related Work}

Previous work has used textual analysis of Twitter posts to study diverse and global populations, including investigating temporal changes in mood~\cite{golder2011diurnal} and correlations between religious expression and sentiment~\cite{ritter2013happy}.
Several other works predict latent characteristics of individuals and communities using social media posts and metadata. 
Rao et al.~\cite{rao2010classifying} predict gender, age, regional origin, and political orientation for individual Twitter users, using tweets and a set of hand-constructed linguistic features.
Burger et al.~\cite{burger2011discriminating} and Bamman et al.~\cite{bamman2014gender} predict users' gender using their tweets and additional meta information, such as name, self description, and their social network.
Jurafsky et al.~\cite{jurafsky14} analyze a corpus of restaurant reviews and predict restaurant ratings using linguistic features such as sentiment, narrative, and self-portrayal.

Paul and Dredze~\cite{paul2011you} apply the Ailment Topic Aspect Model to 1.5 million health-related tweets and discover mentions of over a dozen ailments, including allergies and insomnia. 
Schwartz et al.~\cite{schwartz2013characterizing} use Twitter to predict public health and well-being statistics on a state-wide level. The language used in 82 million geo-located tweets from 1,300 different US
counties is used to predict the subjective well-being of people living in those counties. As in our study, LDA topics improved accuracy.
Hingle et al.~\cite{hingle2013} use Twitter together with analytical software to capture real-time food consumption and diet-related behavior. While this study identifies relationships between dietary and behavioral patterns the results were based on a small dataset (50 participants and 773 tweets).
Nascimento et al.~\cite{jmir3265} evaluate self-reported migraine headache suffering using over 20,000 migrane-related tweets over a seven-day period, finding different peaking hours on weekdays and weekends.
Yom-Tov et al.~\cite{jmir.3156} show how Twitter can be used to discover possible outbreaks of communicable diseases at large public gatherings. Mysl{\'i}n et al.~\cite{jmir2534} use machine classification of tobacco-related Twitter posts to detect tobacco-relevant posts and sentiment towards tobacco products.

Previous work has modelled linguistic variation on Twitter in terms of demographic and geographic variables. O'Connor et al.~\cite{o2010mixture} create a generative model of word use from demographic traits, and show clusters of Twitter users with common lexicons.
Eisenstein et al.~\cite{eisenstein2010latent,eisenstein2012mapping} show that despite the global diffusion of social media, geographic regions have distinct word and topic use on Twitter. 

Previous work has also developed systems for aggregating, processing, and visualizing tweets.
McCreadie et al.~\cite{mccreadie2013scalable} develop a system for detecting newsworthy events and clustering tweets in real-time.
Nguyen et al.~\cite{nguyen2013royal} produce geographic visualizations of tweet sentiment using machine learning classifiers and the tweets' location metadata.

Our work builds upon these previous results, and is, to our knowledge, the first to provide a large-scale, empirical analysis of the predictive power of the language of food.

\section{Conclusion and Future Work}

This work empirically demonstrates that food and food discussion are a major part of who we are.
We develop a system for collecting a large corpus of food-related tweets and using these tweets to predict many latent population characteristics: overweight and diabetes rates, political learning, and geographic location of authors.
Furthermore, we integrate several visualization tools that summarize and query this data, allowing us to discover more complex geographical/temporal trends that are driven by the language of food, such as potential migration patterns.
Our analysis indicates that the language of food alone is extremely powerful. For example, on most predictive tasks, a closed vocabulary of only 800 food words approaches the peak performance obtained when using the entire tweets. 
Perhaps most importantly, our analysis of the learned predictive models provides big-data-driven insights into connections between the language of food and the investigated population characteristics.

We note that our choice of populations (e.g., cities, states) for these  tasks is purely practical (driven by the size of Twitter data at this granularity, and availability of dependent variables for the predictive tasks) and not a limitation of the proposed approach. 
In the future we would like to use our system
to predict characteristics of individuals (e.g., propensity for diabetes), using the individuals' food information. 
Given sufficient amounts of available data, this can lead to non-trivial public health applications and, in a commercial and/or political space, to improved targeted marketing.

This paper is accompanied by a supplemental website, 
\url{https://sites.google.com/site/twitter4food/}, which includes a live version of all visualization tools presented.  
\bibliographystyle{abbrv}
\bibliography{refs}

\newcommand{\noopsort}[1]{} \newcommand{\printfirst}[2]{#1}
  \newcommand{\singleletter}[1]{#1} \newcommand{\switchargs}[2]{#2#1}
\begin{thebibliography}{10}

\bibitem{bamman2014gender}
D.~Bamman, J.~Eisenstein, and T.~Schnoebelen.
\newblock Gender identity and lexical variation in social media.
\newblock {\em Journal of Sociolinguistics}, 18(2):135--160, 2014.

\bibitem{barth14experimental}
L.~Barth, S.~G. Kobourov, and S.~Pupyrev.
\newblock Experimental comparison of semantic word clouds.
\newblock In {\em SEA}, pages 247--258, 2014.

\bibitem{blei2003latent}
D.~M. Blei, A.~Y. Ng, and M.~I. Jordan.
\newblock Latent dirichlet allocation.
\newblock {\em The Journal of Machine Learning Research}, 3:993--1022, 2003.

\bibitem{burger2011discriminating}
J.~D. Burger, J.~Henderson, G.~Kim, and G.~Zarrella.
\newblock Discriminating gender on twitter.
\newblock In {\em EMNLP}, pages 1301--1309. Association for Computational
  Linguistics, 2011.

\bibitem{eisenstein2010latent}
J.~Eisenstein, B.~O'Connor, N.~A. Smith, and E.~P. Xing.
\newblock A latent variable model for geographic lexical variation.
\newblock In {\em EMNLP}, pages 1277--1287. Association for Computational
  Linguistics, 2010.

\bibitem{eisenstein2012mapping}
J.~Eisenstein, B.~O'Connor, N.~A. Smith, and E.~P. Xing.
\newblock Mapping the geographical diffusion of new words.
\newblock {\em arXiv preprint arXiv:1210.5268}, 2012.

\bibitem{golder2011diurnal}
S.~A. Golder and M.~W. Macy.
\newblock Diurnal and seasonal mood vary with work, sleep, and daylength across
  diverse cultures.
\newblock {\em Science}, 333(6051):1878--1881, 2011.

\bibitem{hingle2013}
M.~Hingle, D.~Yoon, J.~F. S.~G. Kobourov, M.~Schneider, D.~Falk, and R.~Burd.
\newblock Collection and visualization of dietary behavior and reasons for
  eating using a popular and free social media software application.
\newblock {\em Journal of Medical Internet Research (JMIR)}, 15(6):125--145,
  2013.

\bibitem{jurafsky14}
D.~Jurafsky, V.~Chahuneau, B.~Routledge, and N.~Smith.
\newblock Narrative framing of consumer sentiment in online restaurant reviews.
\newblock {\em First Monday}, 19(4), 2014.

\bibitem{manning2008iir}
C.~D. Manning, P.~Raghavan, and H.~Sch\"{u}tze.
\newblock {\em Introduction to Information Retrieval}.
\newblock Cambridge University Press, New York, NY, USA, 2008.

\bibitem{mccreadie2013scalable}
R.~McCreadie, C.~Macdonald, I.~Ounis, M.~Osborne, and S.~Petrovic.
\newblock Scalable distributed event detection for twitter.
\newblock In {\em Int. Conf. on Big Data, 2013}, pages 543--549. IEEE, 2013.

\bibitem{jmir2534}
M.~Mysl{\'i}n, S.-H. Zhu, W.~Chapman, and M.~Conway.
\newblock Using twitter to examine smoking behavior and perceptions of emerging
  tobacco products.
\newblock {\em J Med Internet Res}, 15(8):e174, Aug 2013.

\bibitem{jmir3265}
D.~T. Nascimento, F.~M. DosSantos, T.~Danciu, M.~DeBoer, H.~van Holsbeeck,
  R.~S. Lucas, C.~Aiello, L.~Khatib, A.~M. Bender, , J.-K. Zubieta, and F.~A.
  DaSilva.
\newblock Real-time sharing and expression of migraine headache suffering on
  {Twitter}: A cross-sectional infodemiology study.
\newblock {\em J Med Internet Res}, 16(4):e96, Apr 2014.

\bibitem{nguyen2013royal}
V.~D. Nguyen, B.~Varghese, and A.~Barker.
\newblock The royal birth of 2013: Analysing and visualising public sentiment
  in the uk using twitter.
\newblock In {\em Int. Conf. on Big Data, 2013}, pages 46--54. IEEE, 2013.

\bibitem{o2010mixture}
B.~O’Connor, J.~Eisenstein, E.~P. Xing, and N.~A. Smith.
\newblock A mixture model of demographic lexical variation.
\newblock In {\em Proc. of NIPS workshop on machine learning in computational
  social science}, pages 1--7, 2010.

\bibitem{paul2011you}
M.~J. Paul and M.~Dredze.
\newblock You are what you tweet: Analyzing {Twitter} for public health.
\newblock In {\em ICWSM}, 2011.

\bibitem{ranganath2009s}
R.~Ranganath, D.~Jurafsky, and D.~McFarland.
\newblock It's not you, it's me: Detecting flirting and its misperception in
  speed-dates.
\newblock In {\em EMNLP}, pages 334--342, 2009.

\bibitem{rao2010classifying}
D.~Rao, D.~Yarowsky, A.~Shreevats, and M.~Gupta.
\newblock Classifying latent user attributes in {Twitter}.
\newblock In {\em 2nd Intl.~Workshop on Search and mining user-generated
  contents}, pages 37--44. ACM, 2010.

\bibitem{ritter2013happy}
R.~S. Ritter, J.~L. Preston, and I.~Hernandez.
\newblock Happy tweets: Christians are happier, more socially connected, and
  less analytical than atheists on {Twitter}.
\newblock {\em Social Psychological and Personality Science}, pages 243--249,
  2013.

\bibitem{schwartz2013characterizing}
H.~Schwartz, J.~Eichstaedt, M.~Kern, L.~Dziurzynski, M.~Agrawal, G.~Park,
  S.~Lakshmikanth, S.~Jha, M.~Seligman, L.~Ungar, et~al.
\newblock Characterizing geographic variation in well-being using tweets.
\newblock In {\em 7th Intl. AAAI ICWSM}, 2013.

\bibitem{socher2013}
R.~Socher, A.~Perelygin, J.~Wu, J.~Chuang, C.~D. Manning, A.~Ng, and C.~Potts.
\newblock Recursive deep models for semantic compositionality over a sentiment
  treebank.
\newblock In {\em EMNLP}. Association for Computational Linguistics, 2013.

\bibitem{vapnik1998}
V.~N. Vapnik.
\newblock {\em Statistical Learning Theory}.
\newblock John Wiley and Sons, Inc., New York, NY, USA, 1998.

\bibitem{jmir.3156}
E.~Yom-Tov, D.~Borsa, J.~I. Cox, and A.~R. McKendry.
\newblock Detecting disease outbreaks in mass gatherings using internet data.
\newblock {\em J Med Internet Res}, 16(6):e154, Jun 2014.

\end{thebibliography}
\end{document}